\documentclass[lettersize,journal]{IEEEtran}
\usepackage{amsmath,amsfonts}
\usepackage{algorithmic}
\usepackage{algorithm}
\usepackage{array}
\usepackage[caption=false,font=normalsize,labelfont=sf,textfont=sf]{subfig}
\usepackage{textcomp}
\usepackage{stfloats}
\usepackage{url}
\usepackage{verbatim}
\usepackage{graphicx}
\usepackage{cite}
\usepackage{multirow} 
\usepackage{caption}
\usepackage{graphicx}
\usepackage{array}
\usepackage{import}
\usepackage{xcolor}
\usepackage{colortbl}
\usepackage{booktabs}
\usepackage{orcidlink} 
\def\BibTeX{{\rm B\kern-.05em{\sc i\kern-.025em b}\kern-.08em
    T\kern-.1667em\lower.7ex\hbox{E}\kern-.125emX}}
\usepackage{balance}
\begin{document}
\title{GazeCLIP: Gaze-Guided CLIP with Adaptive-Enhanced Fine-Grained Language Prompt for Deepfake Attribution and Detection}
\author{Yaning Zhang \orcidlink{0000-0001-8442-2777},   Linlin Shen \orcidlink{0000-0003-1420-0815}, \emph{Senior Member, IEEE}, Zitong Yu \orcidlink{0000-0003-0422-6616}, \emph{Senior Member, IEEE}, Chunjie Ma \orcidlink{0000-0002-6348-671X}, and Zan Gao \orcidlink{0000-0003-2182-5741}, \emph{Senior Member, IEEE}
	
\thanks{This work was supported in part by the National Natural Science Foundation of China (No.U25A20444, No.62372325, No.62402255), Natural Science Foundation of Tianjin Municipality (No.23JCZDJC00280), Shandong Provincial Natural Science Foundation (No.ZR2024QF020), in part by Shandong Provincial
	Natural Science Foundation under Grant ZR2024QF020 and ZR2026ZD51, Shandong Province National Talents Supporting Program (No.2023GJJLJRC-070), Shandong project towards the integration of education and industry (No.801822020100000024), Young Talent of Lifting engineering for Science and Technology in Shandong (No. SDAST2024QTB001), Shandong Project towards the Integration of Education and Industry (No.2024ZDZX11) (Corresponding author: Zan Gao)}

\thanks{Y. Zhang is with Faculty of Computer Science and Technology, Qilu University of Technology (Shandong Academy of Sciences), Jinan, 250014, China. E-mail: zhangyaning0321@163.com}

\thanks{L. Shen is with Computer Vision Institute, College of Computer Science and Software Engineering, Shenzhen University, Shenzhen, 518060, China, also with National Engineering Laboratory for Big Data System Computing Technology, Shenzhen University, also with Shenzhen Institute of Artificial Intelligence and Robotics for Society, Shenzhen, 518129 and also with Guangdong Key Laboratory of Intelligent Information Processing, Shenzhen University. E-mail: llshen@szu.edu.cn}

\thanks{Z. Yu is with School of Computing and Information Technology, Great Bay University, Dongguan, 523000, China. E-mail: yuzitong@gbu.edu.cn}

 \thanks{C. Ma is with the Shandong Artificial Intelligence Institute, Qilu University of
	Technology (Shandong Academy of Sciences), Jinan, 250014, China. E-mail: mcj@machunjie.com }

\thanks{Z. Gao is with the Shandong Artificial Intelligence Institute, Qilu University of Technology (Shandong Academy of Sciences), Jinan, 250014, China, and also with the Key Laboratory of Computer Vision and System, Ministry of Education, Tianjin University of Technology, Tianjin, 300384, China. E-mail: zangaonsh4522@gmail.com }

}
\markboth{Journal of \LaTeX\ Class Files,~Vol.~18, No.~9, September~2020}%
{How to Use the IEEEtran \LaTeX \ Templates}
\maketitle

\begin{abstract}
	The challenge of tracing the source attribution and authenticity of forged faces has attracted significant attention. Current deepfake attribution or deepfake detection works tend to exhibit poor generalization to novel generative methods due to the limited exploration in visual modalities alone. They tend to assess the attribution or detection performance of models on unseen advanced generators, coarsely, and fail to consider the synergy of the two tasks. To this end, we propose a novel gaze-guided CLIP with adaptive-enhanced fine-grained language prompts for fine-grained deepfake attribution and detection (DFAD). Specifically, we conduct a novel and fine-grained benchmark to evaluate the  DFAD performance of networks on novel generators like diffusion and flow models. Additionally, we introduce a gaze-aware model based on CLIP, namely GazeCLIP, which enhances the generalization to unseen face forgery attacks. Built upon the novel observation of significant distribution differences between pristine and forged gaze embeddings, and the preservation distinction of the target gaze in GAN and diffusion-synthesized facial images, we design a visual perception encoder (VPE) to employ the inherent gaze differences to mine global forgery embeddings across appearance and gaze domains. We propose a gaze-aware image encoder (GIE) that fuses forgery gaze prompts extracted via a gaze encoder with common forged image embeddings to capture general attribution patterns, allowing features to be transformed into a more stable and common DFAD feature space. We build a language refinement encoder (LRE) to generate dynamically enhanced language embeddings via an adaptive-enhanced word selector (AWS) for precise vision-language matching. The plug-and-play AWS can be applied to any transformer-based language models with a slight growth in parameters and computational cost. Extensive experiments on our benchmark show that our model outperforms the state-of-the-art by 6.56\% ACC and 5.32\% AUC in average performance under the attribution and detection settings, respectively. Codes are available at \url{https://github.com/Jenine-321/GazeCLIP}.
\end{abstract}


\section{Introduction}
\begin{figure*}[t]
	\centering
	\includegraphics[width=\linewidth]{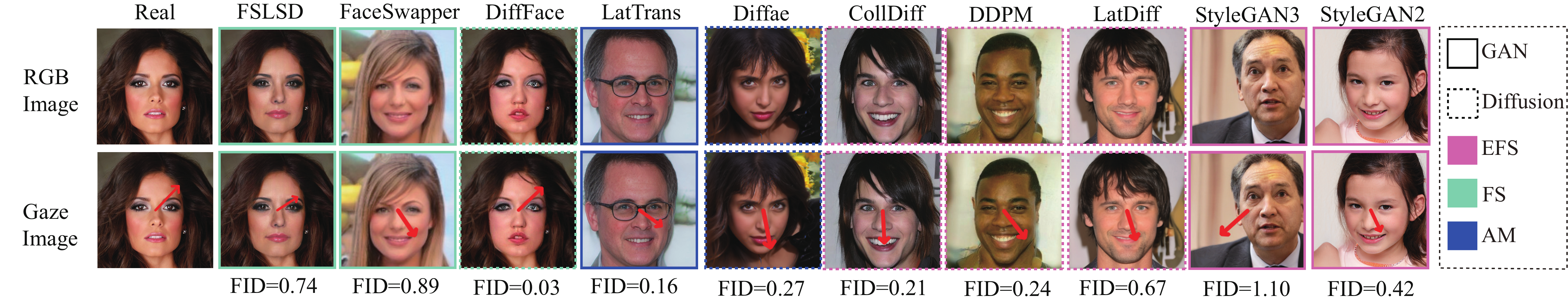}
	\caption{Illustration of the gaze-sensitive prior trait. Each column shows a face yielded by various generators. The first to second rows display the RGB image and the gaze image derived from the pre-trained gaze estimator, respectively. We randomly select 10k real gaze prior embeddings and 10k fake ones for each generator to calculate the FID score. The higher the FID score, the greater the difference between the real and fake gaze embedding distributions. EFS denotes entire face synthesis, FS is face swap, and AM means attribute manipulation.}
	\label{fig1}
\end{figure*}

\begin{figure}[t]
	\centering
	\includegraphics[width=\linewidth]{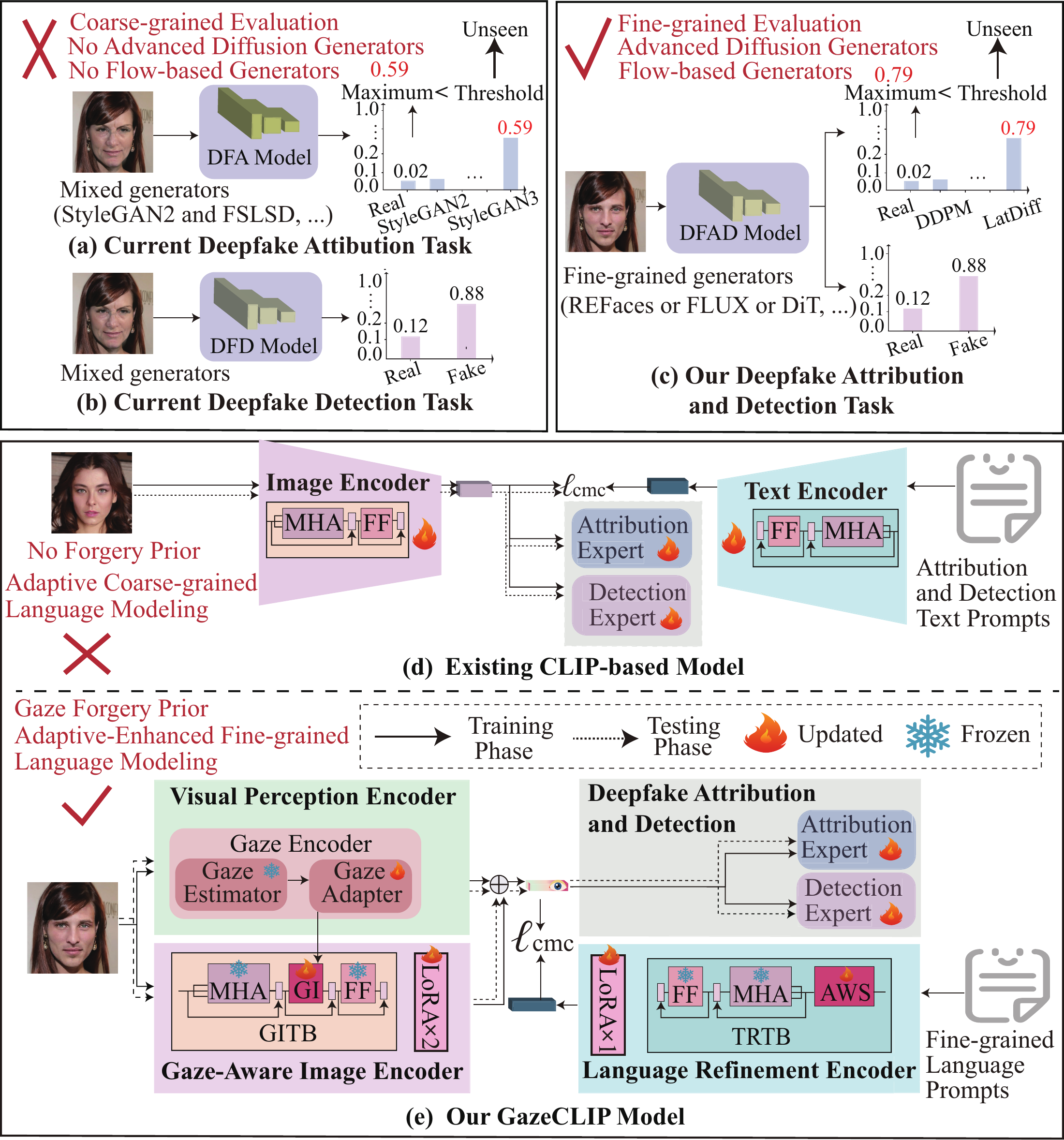}
	\caption{Existing DFA (a) or DFD (b) tasks conduct the coarse-grained evaluation, with generators that are mixed and lack flow models, and fail to achieve collaboration between DFA and DFD. (c) Our DFAD task features the fine-grained evaluation of the model generalization on advanced generators, such as diffusion and flow models, and realize the synergy of DFA and DFD. (d) Existing CLIP-based models hardly introduce the forgery prior and perform dynamically enhanced language modelling. (e) Our GazeCLIP method employs the novel gaze prior and explores the adaptive-enhanced fine-grained language embeddings to achieve powerful generalization.  }
	\label{fig2}
\end{figure}

With the advancement of artificial intelligence generated content like deepfake, sophisticated generative models facilitate seamless face image editing, increasingly blurring the distinction between genuine and forged faces. Deepfake enables the manipulation of face images using advanced generators like generative adversarial networks (GAN) \cite{gan} or diffusion \cite{diff,coll}, with few detectable traces. While it has benefited filmmaking and virtual reality, deepfake has contributed to a rise in malicious face tampering and unauthorized use. Consequently, the need for deepfake attribution and detection (DFAD) has become both critical and urgent. DFAD aims to trace the source of a deepfake image and detect the authenticity using deep learning-based methods. Current research \cite{MAT,forensics,DEFAKE} primarily focuses on deepfake detection (DFD) or deepfake attribution (DFA) separately, with limited exploration into their integration. Existing DFD \cite{DTN,TransDFD} or DFA works \cite{DNADet,cdal} face some limitations in terms of practicality and generalizability. \textbf{First}, as shown in Figure~\ref{fig2} (a), they tend to fail to evaluate the attribution and detection performance of models on unseen advanced generators like diffusion \cite{diff,coll} or flow models \cite{flux}, fine-grainedly. Some works \cite{DEFAKE,DNADet} mine architecture traces across multiple domains to attribute or detection GAN content under a closed-world setting, which constrains their applicability in open-world contexts where novel forgery attacks are constantly evolving. \textbf{Second}, existing vision-language-based models \cite{clip,MFCLIP, Lin} are inclined to focus only on coarse-grained image-language modalities, and hardly consider the fine-grained adaptive-enhanced language modelling and distinct visual forensics prior. MFCLIP \cite{MFCLIP} combines fine-grained noises with image forgery traces and enhances visual features via static fine-grained language guidance to achieve DFD.

To address the limitation, we introduce a novel DFAD benchmark. Unlike current DFA or DFD tasks (See Figure~\ref{fig2}) that conduct coarse-grained model performance evaluation on unseen GAN generators, our DFAD task aims to trace and detect novel advanced generators, such as diffusion and flow models, fine-grainedly. Specifically, our benchmark encompasses about 20 advanced generators, divided into seen and unseen categories for attribution, and real and fake categories for detection. The challenge of DFAD is how to improve the generalization ability of models. Inspired by the contrastive language-image pre-training (CLIP) \cite{clip} which excels in generalization, we propose a GazeCLIP model to achieve generalizable DFAD. Unlike CLIP-based methods \cite{ForensicsAdapter,MFCLIP} which tend to concentrate on image and text modalities for DFA or DFD, our GazeCLIP introduces adaptively enhanced fine-grained language prompts and the novel gaze prior, to adapt CLIP to context-aware and precise DFAD scenarios, to achieve the collaboration between DFA and DFD. Especially, our GazeCLIP method differs from CLIP-based models in the following aspects (See Figure~\ref{fig2}): \textbf{First}, since most face manipulations such as face swap (FS) struggle to fully preserve facial gaze of the real target image (See Figure~\ref{fig1}), we employ a pre-trained gaze estimator \cite{ETHXGaze} to generate gaze embeddings to focus on fine-grained facial gaze areas. As Figure~\ref{fig1} shows, we observe that there are significant differences between pristine and forged gaze embedding distributions. Specifically, gaze embeddings derived from diffusion-generated faces are more similar to real target gaze ones than those extracted from GAN-synthesized faces, since the FID score of the diffusion-generated gaze mode is lower than that of the GAN-synthesized one. This phenomenon facilitates deepfake detection. Besides, there are evident FID score differences in facial gaze distributions among various generators, which can boost the deepfake attribution. Based on the above observation and considering that different generators may produce various facial gaze manipulation artifacts, we devise a gaze encoder (GE) to focus on high-level, fine-grained, and diverse gaze semantic forgery traces. Furthermore, we design a visual perception encoder (VPE) to explore gaze-aware visual forgery features, adaptively. \textbf{Second}, to fully utilize CLIP world image prior to further improve generalization, we devise a gaze-aware image encoder (GIE) to inject gaze information into the frozen CLIP image encoder via a gaze injector, and adapt it to our DFAD domain using LoRA \cite{lora}, to mine the gaze-aware common forgery patterns. Third, since language guidance mitigates generator-specific shifts \cite{MFCLIP}, we propose a language refinement encoder (LRE) to better exploit CLIP language priors. LRE introduces dynamically enhanced fine-grained language prompts into the frozen CLIP text encoder via an adaptive-enhanced word selector (AWS), and adapts them to our text domain using LoRA for precise vision-language alignment. This improves the generalizable attribution ability of the model. In summary, the contributions of this work are as follows:

$\bullet$ To the best of our knowledge, we conduct the first fine-grained deepfake attribution and detection (DFAD) benchmark, which involves advanced generators like diffusion and flow models, to boost the DFAD application in open-world scenarios where novel deepfake attacks emerge.

$\bullet$ Based on novel findings of high-level gaze feature differences, we propose a GazeCLIP model, which combines gaze-aware forgery features with general CLIP image priors and explores adaptive-enhanced fine-grained language embeddings for vision-language matching, thus improving the generalization to unseen generators.  

$\bullet$ We design an innovative plug-and-play adaptive-enhanced word selector to flexibly focus on relevant language features and suppress irrelevant ones, which could be integrated into any transformer-based language models with only a slight increase in parameters and computational costs.

$\bullet$ Extensive experiments conducted on our benchmark demonstrate that our GazeCLIP method outperforms the state of the art under the DFAD scenarios, and deepfake attribution tasks tend to boost the deepfake detection.

\section{Related Work}
\label{sec:formatting}
\subsection{Deepfake Attribution}
Deepfake attribution aims to identify the source of deepfake through deep learning-based models. DNA-Det \cite{DNADet} captures globally consistent generator traces by pre-training on image transformation classification and patch-based contrastive learning. DE-FAKE \cite{DEFAKE} uses the CLIP model to attribute fake images generated by diffusion-based text-to-image generation models. OmniDFA \cite{omnidfa} uses a dual-path framework to capture both fine-grained and global image features, enhancing feature discrimination with contrastive learning and clustering real samples via sphere center loss, for open-set synthesis image detection and few-shot attribution. CDAL \cite{cdal} models the causal relationships between visual attention traces and source attribution, decoupling attribution artifacts from source biases. VLForgery \cite{vlforgery} proposes a knowledge-guided description method that leverages image-generation knowledge and low-level real–fake comparisons to enable efficient, structured forgery analysis from detection to localization and attribution. By contrast, we leverage rarely considered gaze priors and adaptively enhanced fine-grained language prompts to guide CLIP to learn general image forgery patterns, to realize the synergy between deepfake detection and attribution tasks.

\subsection{Deepfake Detection}
Existing deep learning-based methods have been proposed to detect the content caused by deepfake. Dang et al. \cite{DFFD} leverage the Xception backbone to capture local forgery traces for deepfake detection, which rarely considers global information. To address this problem, Wodajo et al. \cite{CViT} propose a convolutional vision transformer (CViT) that combines CNN with the vision transformer (ViT) to mine local and global manipulated traces. 
CAEL \cite{genface} mines multi-grained appearance and edge global forgery traces, and explores the diverse fusion across two domains to mine complementary artifacts. ForensicAdapter \cite{forensics} develops an adapter to identify the distinctive blending boundaries in forged faces, and further improves the visual tokens of CLIP by employing a tailored interaction strategy, to boost the transfer of knowledge between CLIP and the adapter. To improve the generalization to unseen forgery attacks, some works explore the vision language models (VLM) to identify the comprehensive and diverse manipulated patterns. Xu et al. \cite{IVLM} propose a lightweight personalized VLM that dynamically learns identity priors via encoding appearance and behaviour tokens, which incorporates a detection adapter to extract low-level artifact from shallow vision encoder layers and integrate them with high-level identity semantic embeddings for explainable face forgery detection. Chen et al. \cite{componentVLM} propose a VLM-based DFD framework that formulates detection as a VQA task and employs component-specific prompts to focus on forgery-sensitive facial regions. MFVLR \cite{mfvlr} captures general and complementary visual forgery patterns across the image and residual domains and enhances them via fine-grained vision-language matching, to achieve generalizable diffusion-synthesized face forgery detection and localization.

Unlike existing VLM-based methods for DFA or DFD, which primarily rely on generic image-text alignment, our GazeCLIP model explicitly introduces gaze-derived forensic priors to guide visual representation learning toward forgery-sensitive regions via GIE. This provides task-specific forensic information that is largely absent from standard semantic alignment, enabling the model to capture subtle and generalizable gaze-related forgery cues. Second, GazeCLIP provides dynamic forensic guidance via LRE rather than relying on static or generic textual descriptions. By adaptively refining the language representation according to fine-grained visual cues, the language encoder can better emphasize discriminative forgery attributes and establish more precise vision-language alignment. Importantly, these two components play complementary roles: GIE determines where to attend and which visual forensic cues to capture, while LRE determines how to semantically characterize and enhance these cues. Their interaction allows GazeCLIP to move beyond generic image-text alignment toward gaze-guided, fine-grained forensic representation learning.
\subsection{Vision-Language Models}
Vision-language models \cite{clip,Lin,ForensicsAdapter} such as CLIP \cite{clip}, jointly train an image encoder and a text encoder to align image-text pairs from the training datasets. CLIP enables inference on downstream tasks using a zero-shot linear classifier, which is equipped with class names or descriptions from the target dataset. To adjust CLIP to the DFD domain, ForensicsAdapter \cite{ForensicsAdapter} introduces an adapter that digs unique blending boundaries of forged faces, and then enhances CLIP visual tokens with a specialized interaction strategy that facilitates knowledge transfer between CLIP and the adapter. MFCLIP \cite{MFCLIP} combines fine-grained noises from the richest patches with global image forgery artifacts and enhances visual features across image-noise modalities through adaptive vision-language matching. Lin et al. \cite{Lin} use learnable visual perturbations to refine feature extraction for DFD, and then leverage face embeddings to generate sample-level adaptive text prompts. By contrast, based on the novel finding of high-level gaze semantic feature differences between GAN and diffusion models, we employ general gaze embeddings and adaptive-enhanced fine-grained text prompts to boost the gaze-aware visual forgery representation learning, to realize fine-grained DFAD.

\begin{figure*}[t!]
	\centering
	\includegraphics[width=\linewidth]{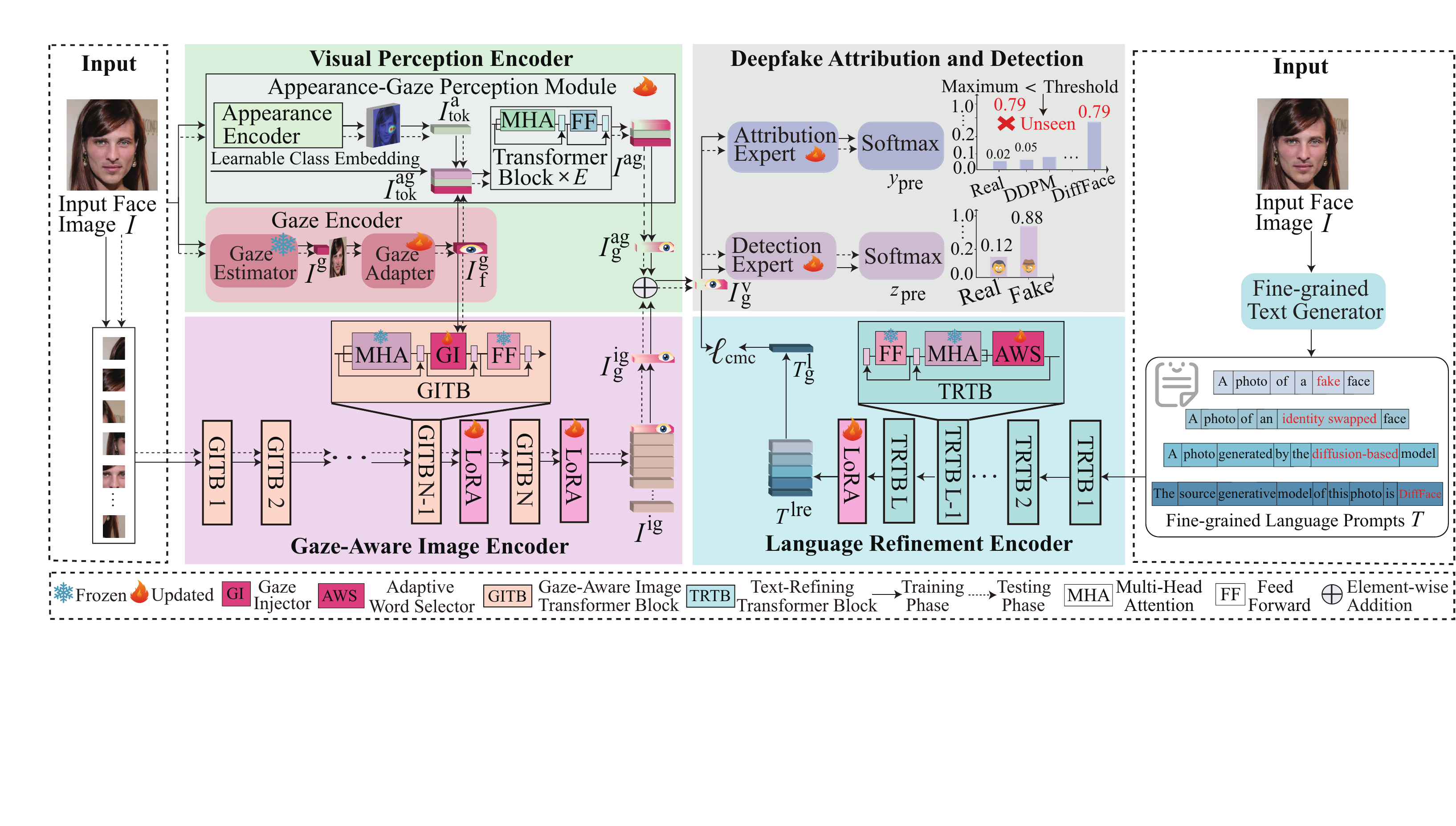} 
	\vspace{-8em}
	\caption{The workflow of our proposed GazeCLIP. After obtaining multiple patches and fine-grained texts of the input face image, VPE is employed to generate gaze features via the gaze encoder and global gaze-aware appearance forgery patterns by AGPM. Then the gaze features and multiple patches are passed into GIE to generate general gaze-guided image embeddings. They are then fused with gaze-appearance forgery traces to derive visual counterfeit features. LRE generates adaptively enhanced language representations via AWS for vision-language matching. At last, the DFAD module takes visual counterfeit embeddings as input to make predictions. During testing, the trained VPE and GIE module are applied to achieve DFAD.  }
	\label{fig3}
\end{figure*}

\section{Methodology}

\subsection{Problem Definition}

For DFAD, given a seen domain $(X_\text{s},Y_\text{s},Z_\text{s})$ and an unseen domain $(X_\text{u},Y_\text{u},Z_\text{u})$, the input distributions of the seen and unseen domains are distinct, and there is no intersection between their attribution label spaces but detection ones, i.e. $X_\text{s} \neq X_\text{u}$, $Y_\text{s}\cap Y_\text{u} = \emptyset$, and  $Z_\text{s} =  Z_\text{u}$, where $X$ denotes the input distribution, $Y$ is the attribution label space, and $Z$ is the detection label space. DFAD aims to learn an attribution and detection model, trained on the seen domain, to trace and identify the distribution from the unseen domain. In our work, we employ the training set $(I_\text{s}$, $y_\text{s}$, $ z_\text{s}) \in \mathcal{D}_\text{tra}$ from the seen domain to train our model, and test it using the testing set $(I_\text{u}$, $y_\text{u}$, $ z_\text{u}) \in \mathcal{D}_\text{tes}$ from the unseen domain, where $I$ denotes the input image, $y \in \{\mathbf{e}_i\mid\mathbf{e}_i=(0,0,\ldots,1,\ldots,0),1\leq i\leq x\} $ denotes generator-specific one-hot labels, $z \in \{[0,1]^T,[1,0]^T\} $ is the real or fake one-hot label, $\mathbf{e}_i \in\mathbb{R}^{x}$ denotes the $i$-th one-hot label, and $x$ is the number of generators. 

\subsection{Method Overview}

Different from previous methods \cite{clip,Lin,ForensicsAdapter} that pay attention to interactions within image or image-language modalities for DFA or DFD, our GazeCLIP model innovatively integrates image, language, and eye gaze modalities to enable DFAD. As Figure~\ref{fig3} shows, our method contains three modules: visual perception encoder (VPE), gaze-aware image encoder (GIE), and language refinement encoder (LRE). 

During training, given an input facial image $I$, we divide it into multiple square image patches, and employ the fine-grained text generator (FTG) to derive language prompts $T$. $I$ and $T$ are then fed into VPE and LRE to derive diverse global forgery embeddings $I_\text{g}^\text{ag}$ across appearance and gaze domains as well as adaptively enhanced fine-grained global language features $T_\text{g}^\text{l}$, accordingly. We send square patches to GIE to gain uniform gaze-perceptual image counterfeiting prompts $I_\text{g}^\text{ig}$. $I_\text{g}^\text{ag}$ and $I_\text{g}^\text{ig}$ are fused to derive  global visual forgery traces $I_\text{g}^\text{v}$. Thereafter, we employ $I_\text{g}^\text{v}$ and $T_\text{g}^\text{l}$ to conduct the language-guided contrastive learning. Meanwhile, we send $I_\text{g}^\text{v}$ to the attribution expert and detection expert in the DFAD module, both consisting of fully connected layers, to yield predicted DFA and DFD logits. Finally, we use the softmax function to produce the attribution prediction $y_\text{pre}$ and detection prediction $z_\text{pre}$, respectively. 

During testing, to prevent text information leakage and considering that novel generators with unknown forgery types may lack precise text prompts, GazeCLIP only uses the trained VPE and GIE to dig gaze-aware common forgery features, which are then fed into the learned DFAD to yield predictions.

\subsection{Visual Perception Encoder}

{\bfseries\setlength\parindent{0em} Gaze encoder.} Unlike existing DFA or DFD methods \cite{Freq,DTN,implicit,Lin} that are inclined to introduce discriminative priors such as landmark, identity, or frequency characteristics, we focus on eye gaze features since the target facial gaze in fake images tends to be incompletely preserved. As Figure~\ref{fig1} shows, considering that there are evident target gaze preservation differences between face images created by GAN and diffusion, gaze embeddings derived from diffusion-generated faces are more similar to real target gaze ones than those extracted from GAN-synthesized faces since the FID score of the diffusion-generated gaze mode is lower than that of the GAN-synthesized one. Besides, there are obvious FID score differences in facial gaze distributions among various generators. Therefore, we aim to utilize the discriminative and general high-level gaze forgery features as prior information to extract robust and common attribution and detection patterns, thus facilitating DFAD. To mine discriminative and common gaze forgery patterns, we design the gaze encoder (GE), which consists of a pre-trained gaze estimator ETH-Xgaze \cite{ETHXGaze} and a gaze adapter. In Figure~\ref{fig3}, given an input facial image $I$, following ETH-Xgaze, we conduct the data normalization and then utilize the frozen pre-trained ETH-Xgaze \cite{ETHXGaze} to obtain gaze embeddings $I^\text{g}\in\mathbb{R}^{1\times 2}$. To perform feature alignment and adapt gaze information to our DFAD domain, we propose the gaze adapter with a fully connected layer to derive gaze forgery traces ${I}^\text{g}_\text{f}\in\mathbb{R}^{1\times d}$, i.e., ${I}^\text{g}_\text{f} ={I}^\text{g}W^{\textrm{g}}$, where $W^{\textrm{g}}\in\mathbb{R}^{2 \times d} $ is the trainable weight of the gaze adapter trained from scratch. $I^\text{g}_\text{f}$ is then fed into the AGPM and GIE to conduct gaze-guided face forgery representation learning, respectively.

{\bfseries\setlength\parindent{0em} Appearance-gaze perception module.} 
Since there are evident texture differences between real and fake images \cite{MAT}, and appearance images provide rich texture details, we propose to integrate appearance features with gaze forgery ones, to capture comprehensive global counterfeit embeddings. Different from current multi-domain based methods that tend to adopt the local prior interaction, our AGPM component is capable of achieving global and diverse integration across appearance and gaze forgery embeddings. As Figure~\ref{fig3} shows, AGPM includes an appearance encoder (AE) and an appearance-gaze transformer encoder (AGTE) with $E$ transformer blocks. Concretely, given $I$ and gaze forgery features $I^\text{g}_\text{f}$, to conduct the feature alignment, AGPM sends $I$ to AE with stacked convolutional layers, to derive the local forgery feature map $I^\text{a}_\text{loc}\in\mathbb{R}^{ c\times h\times w}$, where $c$, $h$, $w$ are the channel, height, and width of the feature map, respectively. $I^\text{a}_\text{loc}$ is then flattened along the channel dimension into a 2D embedding and then projected to appearance feature tokens $I^\text{a}_\text{tok}=\text{Proj}(\text{Flat}(I^\text{a}_\text{loc}))\in\mathbb{R}^{ n\times a}$,  where $n =\frac{chw}{chw}$ and $a$ denote the number and dimension of the feature token, respectively. Thereafter, $I^\text{a}_\text{tok}$ is concatenated with the gaze token $I^\text{g}_\text{f}$ and then appended with a learnable class token $I^\text{c}_\text{tok}\in\mathbb{R}^{ 1\times a}$ to aggregate diverse global features across appearance and gaze domains, which can be formulated as: 
\begin{gather}
	{I}_\text{tok}^\text{ag} = [I^\text{c}_\text{tok}||I^\text{a}_\text{tok}||I^\text{g}_\text{f}]\in\mathbb{R}^{ (n+2)\times a}.
\end{gather}

${I}_\text{tok}^\text{ag}$ is then added with learnable position embeddings $P_\text{e}\in\mathbb{R}^{(n+2)\times a}$ to encode the position information, i.e., $I_1^\text{tra}={I}_\text{tok}^\text{ag} \oplus P_\text{e},$ where $\oplus$ denotes element-wise addition.
Subsequently, it is consecutively fed into $E$ blocks to derive $ I^\text{ag}$. AGPM derives global appearance-gaze fused forgery embeddings $I_\text{g}^\text{ag}\in\mathbb{R}^{1\times a} $ using the class token in $ I^\text{ag}$.

\begin{figure}[t!]
	\centering
	\includegraphics[width=\linewidth]{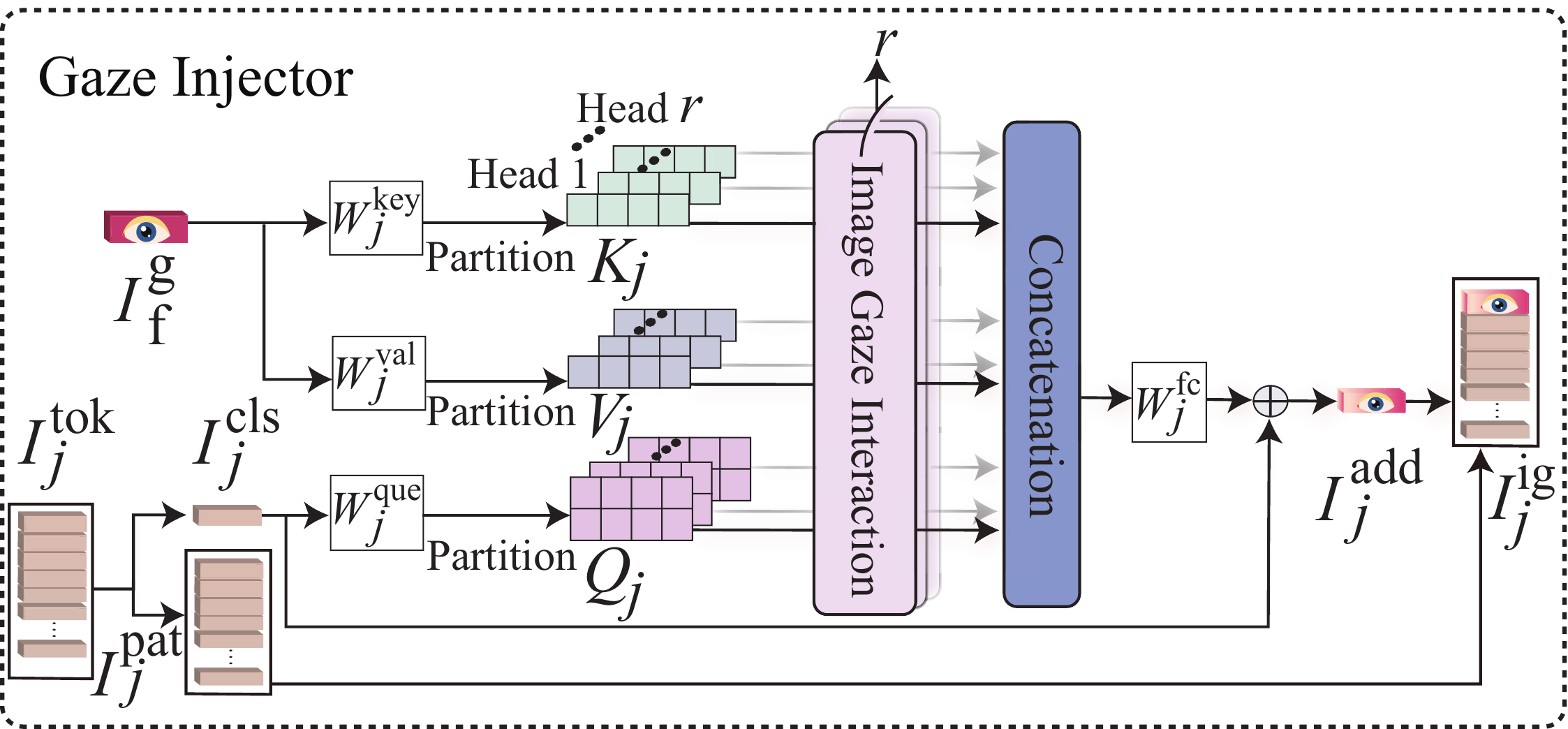}	
	\caption{The pipeline of the gaze injector.  Image forgery embeddings are decomposed into the class token and patch tokens, and only the class token is used as a query to interact with gaze features to extract gaze-aware common forgery patterns.   }\label{fig4}
\end{figure}
\subsection{Gaze-Aware Image Encoder}
Considering that CLIP \cite{clip} is pre-trained using a large-scale publicly available internet dataset of 400 million image-text pairs, it is equipped with general prior knowledge of the world. We aim to integrate general image features encoded by a pre-trained CLIP model with distinct gaze features to explore common visual forgery traces. Unlike vanilla CLIP image encoder that only includes the multi-head attention (MHA) and feed forward (FF) layer in each transformer block, we devise the gaze-aware image encoder (GIE), which adds the proposed gaze injector (GI) after MHA to achieve diverse and global image-gaze interaction. As Figure~\ref{fig3} shows, GIE consists of $N$ gaze-aware image transformer blocks (GITB) $\text{TB}_j^\text{i}$, $j=1,2,…, N$, and each of the last two GITBs is followed by a trainable LoRA \cite{lora} to adjust general image features to our DFAD domain. In detail, given $I$, we divide it into $m$ non-overlapping square patches, and they are then flattened and projected to 2D token sequences $I^\text{i}\in\mathbb{R}^{m\times d}$ with the dimension of $d$ along the channel. Thereafter, $I^\text{i}$ is appended with a learnable class token to learn global forgery features, and then added with a learnable position embedding $P_\text{i}\in\mathbb{R}^{(m+1)\times d}$ to study the position information. That is,
$	I_1^\text{git}=I^\text{i} \oplus P_\text{i}.$ After that, $I_1^\text{git}$ and $I^\text{g}_\text{f}$ are sequentially transmitted into $N$ GITBs and two LoRAs, i.e.,
\begin{align}
	\text{GIE}(I_1^{\text{git}})
	&= \text{LR}_2 \circ \text{TB}^\text{i}_{N}\circ \text{LR}_1 \circ \text{TB}^\text{i}_{N-1} \cdots  \circ\text{TB}_1^\text{i}(I_1^\text{git},I^\text{g}_\text{f} ) \nonumber \\
	&=\text{LR}_2 \circ \text{TB}^\text{i}_{N}\circ \text{LR}_1 \circ \text{TB}^\text{i}_{N-1} (I_{N-1}^\text{git},I^\text{g}_\text{f})
	\nonumber \\
	&= \cdots =\text{LR}_2 \circ \text{TB}^\text{i}_{N}(I_{N}^\text{git},I^\text{g}_\text{f})=I^\text{ig}\nonumber,
\end{align}
where $\circ$ denotes the function decomposition, and LR is LoRA. To fully employ the general world knowledge of the pre-trained CLIP model, each GITB is composed of the frozen pre-trained $\text{MHA}_j^\text{i}$ and $\text{FF}_j^\text{ i}$ layer in the vanilla CLIP image encoder, and a trainable GI module $\text{GI}_j$. Specifically, as Figure~\ref {fig4} illustrates, in the $j$-th GITB ${\text{TB}_j^\text{i}}$, the image forgery embedding $I_{j}^\text{git}$ is first transmitted to the $\text{MHA}_j^\text{i}$ to capture global facial manipulated features $I_j^\text{tok}\in\mathbb{R}^{(m+1)\times d}$. To study abundant and global gaze-aware image forged patterns, we design the gaze injector (GI).   

{\bfseries\setlength\parindent{0em}Gaze injector.} As Figure~\ref{fig4} illustrates, in $\text{GI}_j$, to fuse the image-gaze forged traces more efficiently, we first factorize $I_j^\text{tok}$ into the class token $I_j^\text{cls}\in\mathbb{R}^{1\times d} $ and patch tokens $I_j^\text{pat} \in\mathbb{R}^{m\times d}$. We use only the class token as the query, as it aggregates global facial patch semantics and aligns naturally with the global gaze prior. This design avoids redundant patch-wise gaze interactions while reducing computational cost. The gaze-enhanced class token further propagates gaze information to patch representations through subsequent transformer layers. $\text{GI}_j$ then transforms the $I_j^\text{cls}$ to query matrices using learnable parameter matrices $W^\text{que}_j\in\mathbb{R}^{d\times d}$, and the gaze forgery embedding $I^\text{g}_\text{f}\in\mathbb{R}^{1\times d}$ to key and value matrices via learnable weight matrices $W^\text{key}_j$ and $W^\text{val}_j\in\mathbb{R}^{d\times d}$, respectively.  
\begin{align}
	Q_j &= I_j^\text{cls}W^\text{que}_j, K_j =I^\text{g}_\text{f}W^\text{key}_j, V_j= I^\text{g}_\text{f}W^\text{val}_j.
\end{align}
To realize comprehensive and global relationships between image and gaze forgery features, we perform the image-gaze interaction in parallel,
\begin{align}
	I_{j}^\text{glo} &=\delta(\frac{Q_{j}K_{j}^T}{\sqrt{\frac{r}{d}}})V_{j},
\end{align}

where $\delta$ is the softmax function, and $r$ is the number of heads. Thereafter, it is fed into a FF layer and then added with $I_j^\text{cls}$ to enhance image-gaze counterfeit traces, i.e., $I_{j}^\text{add} = I_{j}^\text{glo}W^\text{fc}_j+I_j^\text{cls}$, where $W^\text{fc}_j\in\mathbb{R}^{d\times d}$ is the learnable weight of the FF layer. It is then concatenated with the patch token $I_j^\text{pat}$ to pass the learned forgery knowledge from the gaze token to its own patch tokens, thereby enhancing the interaction between image patches and gaze, i.e., ${I}_j^\text{ig} = [I_{j}^\text{add}||I_j^\text{pat}]\in\mathbb{R}^{ (m+1)\times d}.$ Finally, ${I}_j^\text{ig}$ is transmitted into a $\text{FF}_j^\text{ i}$ layer with a fully connected layer to yield $I_{j+1}^\text{git}=\text{FF}_j^\text{i}({I}_j^\text{ig})+{I}_j^\text{ig}\in\mathbb{R}^{(m+1)\times d}$. Finally, GIE yields global and common gaze-aware image forgery traces $I_\text{g}^\text{ig}\in\mathbb{R}^{1\times d} $ using the class token in $ I^\text{ig}$. 

\subsection{Language Refinement Encoder}

Like GIE, we intend to employ general text embeddings encoded by a pre-trained CLIP text encoder to explore abundant universal language features. Different from vanilla CLIP text encoder that merely consists of the MHA and FF layer in each transformer block, we devise the language refinement encoder (LRE), which introduces the proposed plug-and-play adaptive-enhanced word selector (AWS) before MHA to refine fine-grained language embeddings, dynamically. As Figure~\ref{fig3} shows, LRE contains $L$ text-refining transformer blocks (TRTB) $\text{TB}_j^\text{t}$, $j=1,2,…, L$, and the last TRTB is followed by a trainable LoRA \cite{lora} to adapt fine-grained embeddings to our DFAD domain. In detail, given $I$, we introduce the FTG \cite{MFCLIP} to build fine-grained and abundant text prompts $T$ based on hierarchical fine-grained labels offered by the GenFace dataset \cite{genface}, i.e., the prompts contain image authenticity, manipulation type, generator type, and generator-specific name descriptions. To safeguard against information leakage from texts, and considering that unseen generators may lack text prompts, we incorporate fine-grained texts to train our network exclusively during training. Specifically, given $T$, LRE first creates a series of word tokens $T^\text{tok}\in\mathbb{R}^{t}$ using the tokenizer \cite{clip}, where $t$ is the number of word tokens. They are then mapped to the text embedding $T^\text{emb}\in\mathbb{R}^{t\times s}$ with the dimension of $s$, and added with learnable position representations $P_\text{l}\in\mathbb{R}^{t\times s}$, i.e., $T_\text{1}^\text{tra}={\ T}^\text{emb} +P_\text{l}$. After that, they are consecutively imparted to $L$ TRTBs and a LoRA, i.e., 
\begin{align}
	\text{LRE}(T_\text{1}^{\text{tra}})
	&=\text{LR}\circ\text{TB}_L^\text{t}\circ\text{TB}_{L-1}^\text{t}\circ \cdots \circ  \text{TB}_2^\text{t} \circ\text{TB}_1^\text{t}(T_1^\text{tra}) \nonumber \\
	&=\text{LR}\circ\text{TB}_L^\text{t}\circ\text{TB}_{L-1}^\text{t}\circ \cdots \circ \text{TB}_2^\text{t}(T_2^\text{tra})
	\nonumber \\
	&= \cdots =\text{LR}\circ\text{TB}_L^\text{t}\circ\text{TB}_{L-1}^\text{t}(T_{L-1}^\text{tra})\nonumber \\
	&= \text{LR}\circ \text{TB}_L^\text{t}(T_L^\text{tra}) = T^\text{lre}.     
\end{align}

To thoroughly utilize the general text knowledge of the pre-trained CLIP model, each TRTB is composed of a trainable AWS module $\text{AWS}_j$ and the frozen pre-trained $\text{MHA}_j^\text{t}$ and $\text{FF}_j^\text{t}$ layer in the vanilla CLIP text encoder. Specifically, as Figure~\ref {fig4} shows, in the $j$-th TRTB ${\text{TB}_j^\text{t}}$, the language embedding $T_{j}^\text{tra}$ is first fed into a plug-and-play AWS module to generate dynamic and fine-grained language features. 

\begin{figure}[t!]
	\centering
	\includegraphics[width=\linewidth]{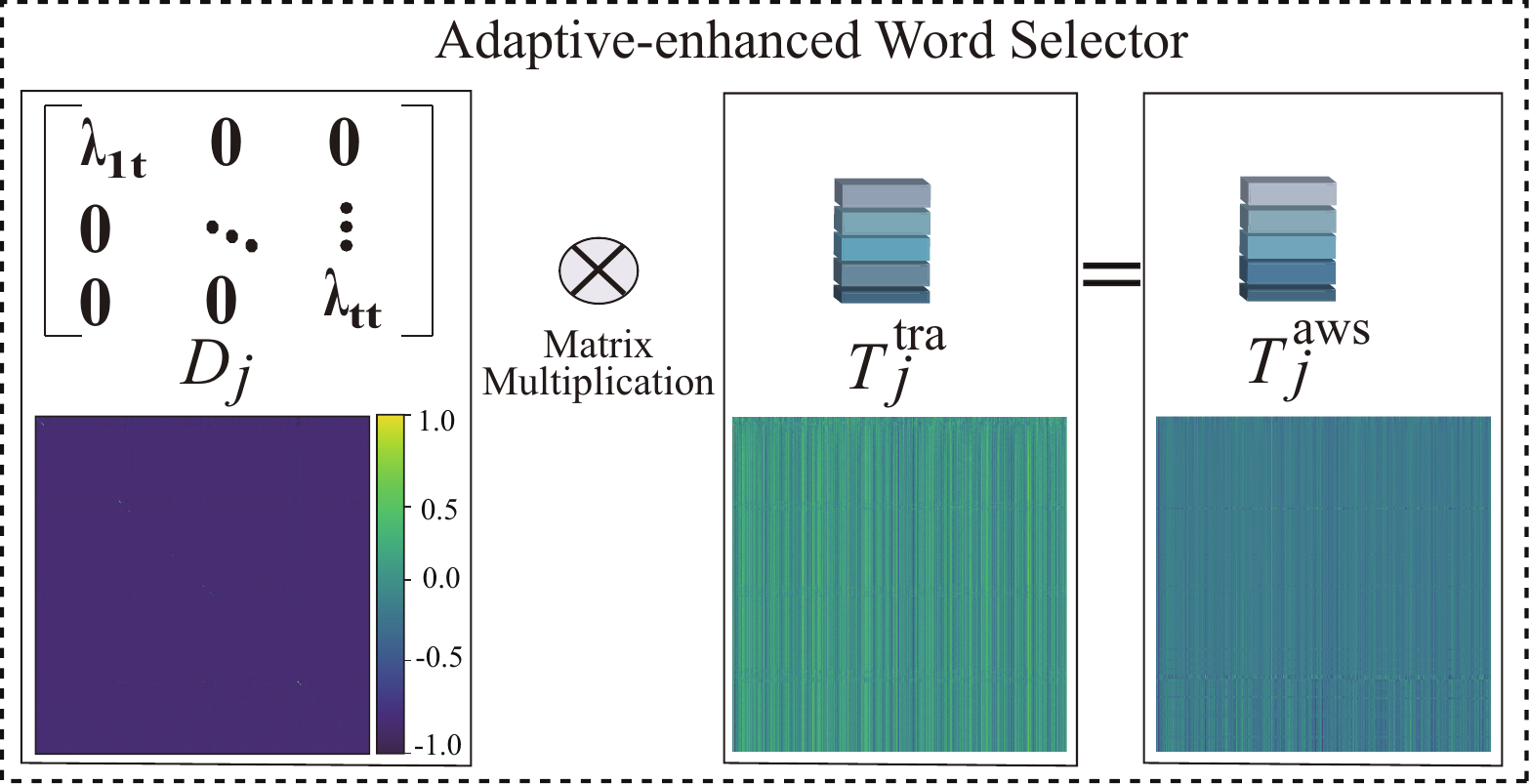}
	\caption{The architecture of the adaptive-enhanced word selector. The brighter (yellow) the color, the more important the word features. }\label{fig5}
\end{figure}

{\bfseries\setlength\parindent{0em} Adaptive-enhanced word selector.} Inspired by the spatial attention mechanism \cite{cbam} in convolutional neural networks (CNN), which assigns learnable weights to each spatial feature to adaptively underline or suppress image patches, we design AWS to dynamically focus on word embeddings in the transformer. AWS study the abundant language representations efficiently and provide a flexible framework that can easily be integrated with different transformer architectures. Unlike existing prompt-based methods that tend to introduce learnable prompts to capture the adaptive relationships among them, our AWS method is designed to achieve adaptively enhanced fine-grained language modelling at the word level, efficiently for DFAD. AWS could be plugged and played into transformer-based language models without requiring extensive architectural changes and parameter count growth.

 Specifically, as Figure~\ref{fig5} shows, in the $j$-th AWS, given normalized fine-grained language embeddings $T_{j}^\text{tra}$, we design a learnable diagonal matrix $D_{j}\in\mathbb{R}^{t\times t}$ to flexibly emphasize important words and suppress unimportant ones. i.e., 
\begin{align}
	T_{j}^\text{aws}
	&=D_{j} T_{j}^\text{tra} \in\mathbb{R}^{t\times s}, \nonumber      
\end{align}
where the elements of the diagonal matrix $D_{j}$ follow the standard normal distribution. We then fed $T_{j}^\text{aws}$ into the frozen pre-trained $\text{MHA}_j^\text{t}$ and $\text{FF}_j^\text{t}$ layer to model global relationships among words, to derive $T_{j+1}^\text{tra}$, which is then imparted to the next TRTB.

 Eventually, LRE generates refined global language features $T_\text{g}^\text{l}\in\mathbb{R}^{1\times s} $ using the last token in $ T^\text{lre}$. $I_\text{g}^\text{ag}$ and $I_\text{g}^\text{ig}$ are mapped and then added element-wisely to derive global visual forgery patterns $I_\text{g}^\text{v}=(I_\text{g}^\text{ig}W_\text{g}^\text{ig}) \oplus (I_\text{g}^\text{ag}W_\text{g}^\text{ag})\in\mathbb{R}^{1\times s},$ where $W_\text{g}^\text{ag}\in\mathbb{R}^{a\times s}$and $W_\text{g}^\text{ig}\in\mathbb{R}^{d\times s}$ are mapping parameters for feature alignment with $T_\text{g}^\text{l}$. We fed $I_\text{g}^\text{v}$ to the attribution expert and detection expert to derive predicted attribution logits $y_\text{pre}\in\mathbb{R}^{1\times x}$ and detection logits $z_\text{pre}\in\mathbb{R}^{1\times 2}$, accordingly.

\subsection{Loss Function}
{\bfseries\setlength\parindent{0em} Deepfake attribution and detection loss.}
 To extract generator-agnostic common forgery features for DFD, we devise the detection loss as follows:
\begin{align}
	\mathcal{L}_\text{dfd}=\ \frac{1}{b}\sum_{u=1}^{b}{-{z^u}^T\text{log}(}z_\text{pre}^u),
\end{align}
where $b$ denotes the number of samples in a batch, and $u \in b$ is the index of samples.
 To explore generator-specific general attribution embeddings for DFA, we devise the attribution loss as follows:
 \begin{align}
 	\mathcal{L}_\text{dfa}=\ \frac{1}{b}\sum_{u=1}^{b}{-{y^u}^T\text{log}(}y_\text{pre}^u).
 \end{align}

{\bfseries\setlength\parindent{0em}Cross-modal contrastive loss.}
Like vanilla CLIP \cite{clip}, we introduce the cross-modal contrastive loss $\mathcal{L}_\text{cmc}$ to conduct the vision-language alignment between $I_\text{g}^\text{ag}$ and $T_\text{g}^\text{l}$.
The total loss is denoted as:
\begin{align}
	\mathcal{L}=\mathcal{L}_\text{dfa}+\mathcal{L}_\text{dfd}+\mathcal{L}_\text{cmc}.
\end{align}

\begin{figure}[t!]
	\centering
	\includegraphics[width=\linewidth]{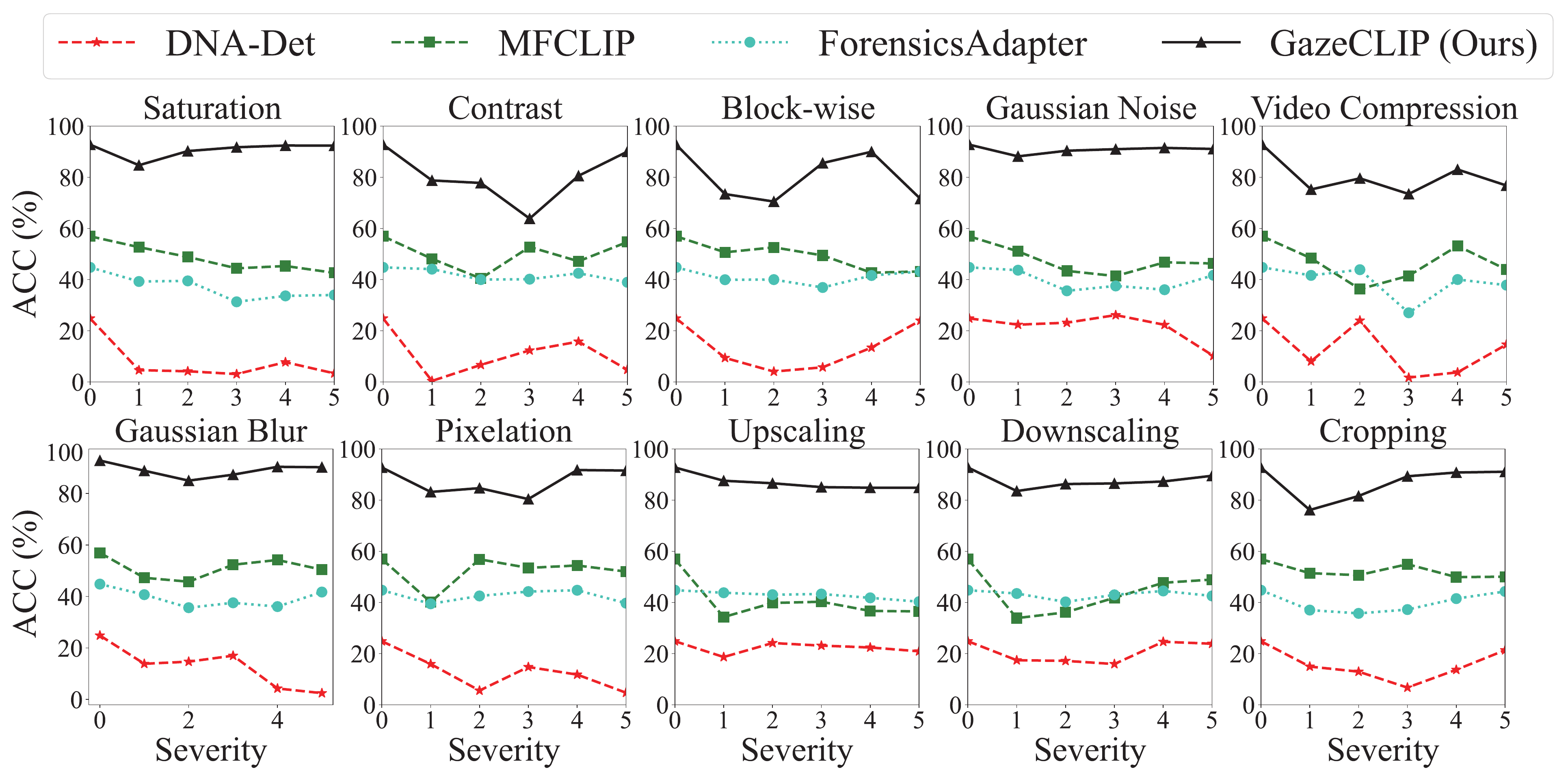} 
	\caption{Resilience to unseen image distortions. }\label{robust2}
\end{figure}

\section{Experiments}

\subsection{Experiment Setup}

{\bfseries\setlength\parindent{0em} Dataset collection and protocol.}
We create real and fake face images generated by various advanced generators, including GAN and diffusion, based on several datasets, such as  GenFace \cite{genface}, DF40 \cite{df40}, CelebA-HQ \cite{celebahq}, Celeb-DF++ \cite{celebdf++}, and FFHQ \cite{ffhq}.  As Table~\ref{tab1} shows, these fake data involve five forgeries, entire face synthesis (EFS), attribute manipulation (AM), face reenactment (FR), talking face (TF), and face swap (FS). Each forgery contains some GAN-based, diffusion-based, and flow-based generators. We offer the detection and attribution label for each generator, accordingly. For deepfake datasets including Celeb-DF, DFDC, and WildDeepfake, the attribution labels are defined at the dataset level based on the generation mechanisms reported in the original dataset papers, rather than independently inferred for each sample. Specifically, Celeb-DF and WildDeepfake are categorized as face-swapping datasets generated using Improved DFAE and DFAE, respectively. Since these generators are unseen during training, both datasets are assigned the unified unseen attribution label $-1$. DFDC contains heterogeneous face-swapping manipulations based on AE, GAN, and MM methods. Therefore, rather than assigning potentially unreliable fine-grained generator labels to individual samples, we treat DFDC as an unseen attribution set and assign the unified label $-1$. We define the protocol to evaluate the attribution and detection performance of various models on unseen generators, deepfake datasets, and FFHQ after training using seen generators and CelebA-HQ.

\begin{table}[t!]
	\centering
	\caption{The outline of the dataset construction and protocol. \# denotes the number of the dataset. 0 means the real image, and 1 is the fake image. -1 denotes the unseen generator. DFAE means deepfake autoencoder. \label{tab1}} 
	\setlength{\tabcolsep}{-0.1mm}{
		\scriptsize
		\begin{tabular}{cccccccrr}
			\toprule
			Seen Sets & Unseen Sets &  Type & Sub-Type & Generator & Train \# & Test \# & \multicolumn{1}{l}{\shortstack{DFD\\Label}} & \multicolumn{1}{l}{\shortstack{DFA\\Label}  } \\
			\midrule
			GenFace &  \multicolumn{1}{c}{-}     & EFS   & GAN   & StyleGAN2 \cite{StyleGAN2} & 10,000 & 1,250 &  \multicolumn{1}{c}{1}   &\multicolumn{1}{c}{9}  \\
			GenFace & \multicolumn{1}{c}{-}      & EFS   & GAN   & StyleGAN3 \cite{StyleGAN3} & 10,000 & 1,250 & \multicolumn{1}{c}{1}   &\multicolumn{1}{c}{10} \\
			GenFace &   \multicolumn{1}{c}{-}    & EFS   & Diffusion & LatDiff \cite{lad} & 10,000 & 1,250 & \multicolumn{1}{c}{1}    &\multicolumn{1}{c}{3}  \\
			GenFace &   \multicolumn{1}{c}{-}    & EFS   & Diffusion & CollDiff \cite{coll} & 10,000 & 1,250 & \multicolumn{1}{c}{1} & \multicolumn{1}{c}{4} \\
			GenFace &    \multicolumn{1}{c}{-}    & EFS   & Diffusion & DDPM \cite{ddpm}  & 10,000 & 1,250 &  \multicolumn{1}{c}{1}     & \multicolumn{1}{c}{5} \\
			GenFace &   \multicolumn{1}{c}{-}     & FS    & GAN   & FSLSD \cite{FSLSD} & 10,000 & 1,250 & \multicolumn{1}{c}{1} &\multicolumn{1}{c}{6} \\
			GenFace &   \multicolumn{1}{c}{-}     & FS    & GAN   & FaceSwapper \cite{FaceSwapper} & 10,000 & 1,250 & \multicolumn{1}{c}{1} &\multicolumn{1}{c}{7}  \\
			GenFace &   \multicolumn{1}{c}{-}     & FS    & Diffusion & DiffFace \cite{difface} & 10,000 & 1,250 & \multicolumn{1}{c}{1} &\multicolumn{1}{c}{1}  \\
			GenFace &   \multicolumn{1}{c}{-}     & AM    & GAN   & LatTrans \cite{Latent} & 10,000 & 1,250 &  \multicolumn{1}{c}{1}   &\multicolumn{1}{c}{8} \\
			GenFace &    \multicolumn{1}{c}{-}    & AM    & Diffusion & Diffae \cite{diffae} & 10,000 & 1,250 &  \multicolumn{1}{c}{1}    &\multicolumn{1}{c}{2}  \\
			\multicolumn{1}{c}{-}	& DF40  & EFS   & GAN   & VQGAN \cite{vqgan} &    \multicolumn{1}{c}{-}   & 1,250 &  \multicolumn{1}{c}{1}   &\multicolumn{1}{c}{-1} \\
			\multicolumn{1}{c}{-}	& DF40  & EFS   & Diffusion & SD-2.1 \cite{sd21} &  \multicolumn{1}{c}{-}     & 1,250 &  \multicolumn{1}{c}{1}   &\multicolumn{1}{c}{-1}  \\
			\multicolumn{1}{c}{-}	& DF40  & EFS   & Diffusion & DiT \cite{DiT}   &  \multicolumn{1}{c}{-}     & 1,250 &  \multicolumn{1}{c}{1} &\multicolumn{1}{c}{-1} \\
			\multicolumn{1}{c}{-}	& DF40  & EFS   & Diffusion & SiT \cite{sit}  &  \multicolumn{1}{c}{-}     & 1,250 &  \multicolumn{1}{c}{1} &\multicolumn{1}{c}{-1} \\
			\multicolumn{1}{c}{-}	& DF40  & FS    & GAN   & SimSwap \cite{simswap}&    \multicolumn{1}{c}{-}   & 1,250 &  \multicolumn{1}{c}{1}   & \multicolumn{1}{c}{-1} \\
			\multicolumn{1}{c}{-}	& DF40  & FS    & GAN   & UniFace \cite{uniface} & \multicolumn{1}{c}{-}      & 1,250 &   \multicolumn{1}{c}{1}    & \multicolumn{1}{c}{-1} \\
			\multicolumn{1}{c}{-}	& DF40  & FS    & GAN   & InSwapper \cite{inswap} &    \multicolumn{1}{c}{-}   & 1,250 & \multicolumn{1}{c}{1}      & \multicolumn{1}{c}{-1} \\
			\multicolumn{1}{c}{-}	& DF40  & FS    & GAN   & e4s \cite{e4s}  &    \multicolumn{1}{c}{-}     & 1,250 &   \multicolumn{1}{c}{1}    & \multicolumn{1}{c}{-1} \\
			\multicolumn{1}{c}{-}	& \multicolumn{1}{c}{-}   & FS    & Diffusion & REFace \cite{refaces} &  \multicolumn{1}{c}{-}     & 1,250 &  \multicolumn{1}{c}{1}   & \multicolumn{1}{c}{-1} \\
			\multicolumn{1}{c}{-}	& Celeb-DF++  & FR    & GAN   & LivePortrait \cite{ liveportrait}  &    \multicolumn{1}{c}{-}     & 1,250 &   \multicolumn{1}{c}{1}    & \multicolumn{1}{c}{-1} \\
			\multicolumn{1}{c}{-}	& Celeb-DF++  & FR    & Flow  & LIA \cite{LIA}  &    \multicolumn{1}{c}{-}     & 1,250 &   \multicolumn{1}{c}{1}    & \multicolumn{1}{c}{-1} \\
			\multicolumn{1}{c}{-}	& Celeb-DF++  & FR    & GAN  & FSRT \cite{fsrt}   &    \multicolumn{1}{c}{-}     & 1,250 &   \multicolumn{1}{c}{1}    & \multicolumn{1}{c}{-1} \\
			\multicolumn{1}{c}{-}	& - & FR   & Diffusion  & DiffusionAct \cite{diffact}    &    \multicolumn{1}{c}{-}     & 1,250 &   \multicolumn{1}{c}{1}    & \multicolumn{1}{c}{-1} \\
			\multicolumn{1}{c}{-}	& Celeb-DF++  & TF    & GAN  & Real3DPortrait \cite{real3d}   &    \multicolumn{1}{c}{-}     & 1,250 &   \multicolumn{1}{c}{1}    & \multicolumn{1}{c}{-1} \\
			\multicolumn{1}{c}{-}	& Celeb-DF++  & TF   & Diffusion  & AniTalker  \cite{anitalker}  &    \multicolumn{1}{c}{-}     & 1,250 &   \multicolumn{1}{c}{1}    & \multicolumn{1}{c}{-1} \\
			\multicolumn{1}{c}{-}	& Celeb-DF++  & TF    & Flow  &FLOAT \cite{float}   &    \multicolumn{1}{c}{-}     & 1,250 &   \multicolumn{1}{c}{1}    & \multicolumn{1}{c}{-1}\\
			\multicolumn{1}{c}{-}	& Celeb-DF++  & TF    & Flow  &EDTalk \cite{edtalk}   &    \multicolumn{1}{c}{-}     & 1,250 &   \multicolumn{1}{c}{1}    & \multicolumn{1}{c}{-1} \\
			\multicolumn{1}{c}{-}	& \multicolumn{1}{c}{-}   & EFS   & Flow & FLUX \cite{flux} &  \multicolumn{1}{c}{-}     & 1,250 &  \multicolumn{1}{c}{1}   & \multicolumn{1}{c}{-1} \\
						\midrule
\multirow{2}{*}{-}
& \multirow{2}{*}{DFDC}
& \multirow{2}{*}{FS}
& \multirow{2}{*}{\shortstack[c]{AE/\\GAN/MM}}
& \multirow{2}{*}{\shortstack[c]{DFAE/MM/NN/NTH/\\FSGAN/StyleGAN}}
& \multirow{2}{*}{-}
& \multirow{2}{*}{1,250}
& \multicolumn{1}{c}{\multirow{2}{*}{1}}
& \multicolumn{1}{c}{\multirow{2}{*}{-1}} \\

& & & & & & & & \\
					\midrule
		-
& Celeb-DF
& FS
& AE
& Improved DFAE
& -
& 1,250
& \multicolumn{1}{c}{1}
& \multicolumn{1}{c}{-1} \\

-
& WildDeepfake
& FS
& AE
& DFAE
& -
& 1,250
& \multicolumn{1}{c}{1}
& \multicolumn{1}{c}{-1} \\
			\bottomrule
		\end{tabular}%
	}

\end{table}

{\bfseries\setlength\parindent{0em}Implementation details.}\label{secA}
We implement the model using PyTorch on a Tesla A100 GPU with 40GB of memory under the Linux system. The number of transformer blocks $E$, $L$, and $N$ in GazeCLIP are set to 6, 12, and 12, respectively. The feature dimensions $a$, $d$, and $s$ are configured to 1024, 768, and 512. The patch number $m$, token number $t$ and $n$, head number $r$, and batch size $b$ are 64, 308, 1, 8, and 32. We set the channel $c$, height $h$, and width $w$ of feature maps to 512, 7, and 7, accordingly. We set the LoRA rank and alpha to 8 and 32. The height and width of images are resized to 256. We conduct the normalization by dividing image pixel values by 255, without any data augmentations. Following ETH-Xgaze \cite{ETHXGaze}, the input face image of the gaze estimator is normalized by mean [0.485, 0.456, 0.406] and standard deviation [0.229, 0.224, 0.225] for each color channel. Our method is trained with the Adam optimizer using a learning rate of 1e-4 and a weight decay of 1e-3. We leverage the scheduler to drop the learning rate by ten times every 15 epochs. To ensure a fair comparison, we develop SOTA methods and follow default configurations until convergence. All models are trained and evaluated using the same protocols. 

{\bfseries\setlength\parindent{0em} Evaluation metrics.}
For DFA, we employ accuracy (ACC) as an evaluation metric. We follow \cite{zhang2025pvlm} to calculate the attribution ACC on unseen generators. We only report results for the threshold of 0.9 in our experiment. For DFD, we use ACC and AUC as metrics.


\subsection{Comparison with the State of the Art}

We evaluate state-of-the-art methods on our benchmark. For DFD models, we add the attribution expert to achieve DFAD. For DFA networks, we introduce the detection expert as the side branch of the attribution expert. We select various deepfake detectors such as CNN-based (ResNet \cite{resnet}, Xception \cite{xception}),  transformer-based (i.e., ViT \cite{ViT}, CViT \cite{CViT}, and CAEL \cite{genface}), vision-language-based (i.e., CLIP \cite{clip}, MFCLIP \cite{MFCLIP}, Lin et al. \cite{Lin}, and ForensicsAdapter (FA) \cite{forensics}), and methods dedicated to attributing deepfake images (e.g., DNA-Det \cite{DNADet}, OmniDFA \cite{omnidfa}, CDAL \cite{cdal}, and DE-FAKE \cite{DEFAKE}).
\begin{table*}[t!]
	\centering  
	\renewcommand{\arraystretch}{1.2}
	\caption{ Comparison of attribution and detection performance across different methods. For attribution, ACC scores (\%) on unseen generators from DF40 and FFHQ, after training using seen generators and CelebA-HQ. For detection, ACC and AUC scores (\%) on unseen generators are averaged. \textbf{Bold} and \underline{underline} denote the best and the second-best results. 	\label{bench}}
	\setlength{\tabcolsep}{1.4mm}{
		\scriptsize
		\begin{tabular}{cccccccccccccc}\toprule\multirow{3}[6]{*}{Model} & \multicolumn{11}{c}{Attribution}                                                      & \multicolumn{2}{c}{ Detection} \\ \cmidrule(lr){2-12} \cmidrule(lr){13-14}    & FFHQ  & SimSwap & InSwapper & UniFace & e4s   & VQGAN & SD-2.1 & DiT   & SiT   & REFace & Average  & \multicolumn{2}{c}{Average} \\ \cmidrule(lr){2-12} \cmidrule(lr){13-14}     & ACC   & ACC   & ACC   & ACC   & ACC   & ACC   & ACC   & ACC   & ACC   & ACC   & ACC   &ACC   & AUC \\ \cmidrule(lr){1-14}ResNet \cite{deep} & 38.24 & 37.44 & 48.80  & 48.00    & 65.36 & 46.24 & 59.60  & 60.56 & 61.44 & 46.48 & 51.22 & 83.10  & 63.22 \\Xception \cite{xception} & 24.64 & 58.08 & 70.64 & 65.28 & 76.16 & 49.04 & \underline{74.32} & \underline{84.32} & \underline{83.52} & 48.64 & 64.02 & 78.58 & 50.78 \\ViT \cite{ViT}   & \underline{48.00}    & 65.52 & 62.00    & 64.24 & 58.48 & 53.12 & 63.92 & 58.72 & 56.80  & 58.32 & 58.91 & 84.42 & 62.31 \\CViT \cite{CViT}  & 1.28  & 34.24 & 36.72 & 31.44 & 49.76 & 13.28 & 49.84 & 66.64 & 65.04 & 36.56 & 38.48 & 86.29 & 64.73 \\CAEL \cite{genface}  & 6.80   & 9.04  & 21.60  & 8.48  & 59.92 & 9.60   & 51.52 & 14.32 & 17.52 & 58.64 & 25.74 & 81.71 & 62.30 \\CLIP \cite{clip}  & \textbf{52.40}  & \underline{78.16} & \underline{88.00}    & \textbf{88.72} & \underline{85.44} & \underline{68.32} & 39.60  & 71.20  & 74.96 & \textbf{94.24} & \underline{74.10}  & 65.55 & 69.91 \\MFCLIP \cite{MFCLIP} & 14.48 & 71.52 & 68.32 & 72.64 & 61.36 & 49.20  & 53.28 & 39.92 & 39.28 & 2.00     & 47.20  & 81.26 & 70.39 \\DNA-Net \cite{DNADet} & 8.16  & 12.24 & 9.52  & 4.72  & 67.04 & 43.04 & 28.40  & 39.20  & 29.60  & \underline{89.52} & 33.14 & 86.99 & \underline{72.03} \\DE-FAKE \cite{DEFAKE} & 22.46  & 1.63 & 1.98  & 2.05  & 18.63 & 4.96& 12.49 & 1.85  & 2.93 & 15.44 & 8.44 & 60.42 & 64.30\\FA \cite{ForensicsAdapter} & 13.84 & 20.64 & 22.56 & 30.40  & 36.24 & 8.24  & 44.24 & 31.52 & 32.88 & 54.00    & 29.46 & \underline{88.78} & 55.45 \\  Lin et al. \cite{Lin}    & 18.96      & 73.20      &    70.85   &  75.61     &   68.32    & 50.79      &  48.16     &    50.67   & 51.14      &    35.70   & 54.28      &   88.04    & 64.80 \\OmniDFA \cite{omnidfa} & 10.26 & 12.78 & 23.64 & 10.51  & 60.32 & 11.49  & 52.64 & 15.37 & 22.45 & 55.72    & 27.52 & 84.36 & 63.18 \\ CDAL \cite{cdal} & 8.65 & 11.87 & 21.09 & 11.38  & 50.36 & 10.67  &42.86 & 13.77 & 20.69 & 53.64    & 24.50 & 80.20 & 53.62 \\ 
			GazeCLIP (Ours) & 5.12  &\textbf{90.16} & \textbf{93.44}   & \underline{79.36}  & \textbf{98.48} & \textbf{96.48} &\textbf{ 98.48} & \textbf{94.00}  & \textbf{92.88} & 58.24 & \textbf{80.66} & \textbf{89.80}  & \textbf{77.35} \\
			\bottomrule
		\end{tabular}%
	}
	
\end{table*}

\begin{table*}[t!]
		\renewcommand{\arraystretch}{1.2}
	\centering  
	\caption{ Comparison of attribution and detection performance across different methods. For attribution, ACC scores (\%) on unseen generators from Celeb-DF++, after training using seen generators and CelebA-HQ. For detection, ACC and AUC scores (\%) on unseen generators are averaged. \textbf{Bold} and \underline{underline} denote the best and the second-best results. 	\label{bench2}}
	\setlength{\tabcolsep}{1.0mm}{
		\scriptsize
		\begin{tabular}{cccccccccccc}\toprule\multirow{3}[6]{*}{Model} & \multicolumn{9}{c}{Attribution}                                                      & \multicolumn{2}{c}{ Detection} \\ \cmidrule(lr){2-10} \cmidrule(lr){11-12}    & LivePortrait  & LIA & FSRT & DiffusionAct & EDTalk   & AniTalker & Real3DPortrait & FLOAT  & Average  & \multicolumn{2}{c}{Average} \\ \cmidrule(lr){2-10} \cmidrule(lr){11-12}     & ACC   & ACC   & ACC   & ACC   & ACC   & ACC   & ACC   & ACC   & ACC   &ACC   & AUC \\ \cmidrule(lr){1-10} \cmidrule(lr){10-12} ResNet \cite{deep} & 42.56 & 40.80 & 46.24  & 62.40    & 51.76 &46.64 & 44.88  & 34.08 & 46.17  &88.74 & \underline{77.95} \\Xception \cite{xception}& 65.28 & 68.16 & 67.28  & \textbf{84.08}    & 40.00 &34.80 & \textbf{97.04}  & 74.80 &66.43 & 85.27  & 35.95\\ViT \cite{ViT}   & 28.16   & 16.88 & 17.84    & 11.20 & 12.80 & 5.92 & 70.48 & \underline{80.56} & 30.48 & 88.87 & 68.72   \\CViT \cite{CViT}  & 27.84  & 16.32 &17.60 & 11.44 & 12.24 & 5.60 & 70.16 & \textbf{80.64} & 30.23 & \underline{88.88} & 68.78  \\CAEL \cite{genface}  & 34.64   & 35.52  & 28.96 & 25.84  & 40.80 & 26.56   & 65.36 &28.40 & 35.76 & 87.53 &66.35   \\CLIP \cite{clip} 
			& \textbf{81.43}
			& \underline{90.64}
			& \underline{92.77}
			& 38.99
			& \underline{94.71}
			& \underline{88.02}
			& 40.92
			& 38.04
			& \underline{70.69}
			& 80.90
			& 71.64 \\MFCLIP \cite{MFCLIP} & 80.56 &89.28 & 91.76 & 36.80 & 92.56 & 85.60  & 28.00 & 19.84 & 65.55 &71.76   & 50.03 \\DNA-Net \cite{DNADet} & 0.00  & 0.00 & 0.00  & 56.00  & 0.00 & 0.00 & 13.92  & 3.20  & 9.14  & 73.49 & 47.06  \\ DE-FAKE \cite{DEFAKE} 
			& 0.00
			& 0.00
			& 0.00
			& 10.75
			& 11.63
			& 3.74
			& 11.78
			& 0.96
			& 4.86
			& 68.03
			& 30.57
			\\FA \cite{ForensicsAdapter} &40.48 & 27.68 & 31.92 & 58.08  & 32.48 & 29.20  & 29.92 & 39.36 & 36.14  & 86.38    &30.92\\ Lin et al. \cite{Lin}
			& 76.08
			& 83.62
			& 85.87
			& 34.79
			& 88.65
			& 83.06
			& 37.35
			& 16.82
			& 62.78
			& 70.94
			& 50.72
			\\
			
			OmniDFA \cite{omnidfa}
			& 5.03
			& 6.82
			& 9.47
			& 12.31
			& 13.98
			& 8.90
			& 19.06
			& 4.24
			& 9.98
			& 72.35
			& 52.88
			\\
			
			CDAL \cite{cdal}
			& 3.80
			& 5.73
			& 7.64
			& 11.85
			& 12.90
			& 7.08
			& 18.76
			& 3.84
			& 8.95
			& 71.06
			& 50.74
			\\
			GazeCLIP (Ours) &\underline{80.08}  &\textbf{92.00} & \textbf{95.04}   & \underline{68.48}  & \textbf{96.40} & \textbf{93.20} &\underline{85.92} & 80.16  & \textbf{86.41} & \textbf{88.89} & \textbf{81.23}  \\
			\bottomrule
		\end{tabular}%
	}	
	
\end{table*}

\begin{table*}[t!]
	\centering  
	\renewcommand{\arraystretch}{1.2}
	\caption{Comparison of forgery type traceability across various methods. ACC scores (\%) on unseen generators from DF40 and FFHQ after training using seen generators and CelebA-HQ. 
			\textbf{Bold} and \underline{underline} are the best and the second-best results.
			\label{ftbench}}
	
	\setlength{\tabcolsep}{1.8mm}{
		\scriptsize
		
		\begin{tabular}{cccccccccccc}
			\toprule
			
			\multirow{3}[6]{*}{Model} 
			& \multicolumn{11}{c}{Forgery type traceability} \\
			
			\cmidrule(lr){2-12}
			
			& FFHQ  
			& SimSwap 
			& InSwapper 
			& UniFace 
			& e4s   
			& VQGAN 
			& SD-2.1 
			& DiT   
			& SiT   
			& REFace 
			& Average \\
			
			\cmidrule(lr){2-12}
			
			& ACC & ACC & ACC & ACC & ACC & ACC 
			& ACC & ACC & ACC & ACC & ACC \\
			
			\cmidrule(lr){1-12}
			
			ResNet \cite{deep} 
			& 5.28 
			& 69.44 
			& 67.92 
			& 53.76 
			& 25.68 
			& 42.16 
			& 70.84 
			& 60.08 
			& 34.64 
			& 18.00 
			& 44.78 \\
			
			Xception \cite{xception} 
			& 3.28 
			& 68.16 
			& 57.04 
			& 49.92 
			& 15.68 
			& 44.48 
			& \underline{98.72} 
			& \underline{76.40} 
			& \underline{76.88} 
			& 13.08 
			& 50.36 \\
			
			ViT \cite{ViT}   
			& 6.88 
			& 42.24 
			& 21.84 
			& 48.40 
			& 25.28 
			& 49.60 
			& 80.80 
			& 68.24 
			& 69.68 
			& 20.32 
			& 43.33 \\
			
			CViT \cite{CViT}
			& 6.64
			& 62.24
			& 54.08
			& 50.00
			& 27.68
			& 14.72
			& 84.24
			& 41.76
			& 42.96
			& 8.76
			& 39.31 \\
			
			CAEL \cite{genface}
			& 10.72
			& \underline{80.40}
			& \underline{69.68}
			& \underline{84.48}
			& 15.84
			& 10.48
			& 84.32
			& 10.48
			& 11.84
			& 9.67
			& 38.79 \\
			
			CLIP \cite{clip}  
			& \textbf{36.78} 
			& 50.44 
			& 48.37 
			& 73.72 
			& 35.80 
			& 50.65 
			& 79.60 
			& 51.20 
			& 74.96 
			& 24.24 
			& \underline{52.58} \\
			
			MFCLIP \cite{MFCLIP} 
			& \underline{33.84} 
			& 29.36 
			& 53.20 
			& 21.28 
			& 10.40 
			& 23.60 
			& 25.20 
			& 27.60 
			& 10.00 
			& 22.48 
			& 25.70 \\
			
			DNA-Net \cite{DNADet} 
			& 28.16 
			& 12.24 
			& 9.52 
			& 4.72 
			& 7.69 
			& 43.04 
			& 28.40 
			& 39.20 
			& 29.60 
			& 9.52 
			& 21.21 \\
			
			DE-FAKE \cite{DEFAKE} 
			& 10.33 
			& 8.76
			& 8.95 
			& 11.06 
			& 12.87 
			& 10.44 
			& 22.95 
			& 20.67 
			& 8.80 
			& 16.89 
			& 13.17 \\
			
			FA \cite{ForensicsAdapter} 
			& 23.84 
			& 30.87 
			& 64.12 
			& 33.96 
			& 40.96 
			& \underline{57.46} 
			& 46.37 
			& 62.74 
			& 53.96 
			& 25.71
			& 44.00 \\
			
			Lin et al. \cite{Lin}    
			& 8.85 
			& 35.96 
			& 40.74 
			& 68.39 
			& 27.94 
			& 51.80 
			& 50.76 
			& 52.43 
			& 51.55 
			& \underline{32.78} 
			& 42.12 \\
			
			OmniDFA \cite{omnidfa} 
			& 10.67 
			& 16.75 
			& 29.35 
			& 20.76 
			& \underline{63.95}
			& 23.57 
			& 55.97 
			& 24.72 
			& 26.19 
			& 20.58 
			& 29.25 \\
			
			CDAL \cite{cdal} 
			& 17.90 
			& 13.88 
			& 24.79 
			& 18.35 
			& 49.45
			& 30.66 
			& 48.58 
			& 27.96 
			& 30.58 
			& 28.67 
			& 29.08 \\
			
			GazeCLIP (Ours) 
			& 7.56 
			& \textbf{93.54} 
			& \textbf{90.86} 
			& \textbf{89.73} 
			& \textbf{70.58} 
			& \textbf{89.50} 
			& \textbf{98.96} 
			& \textbf{85.09} 
			& \textbf{80.93} 
			& \textbf{55.24} 
			& \textbf{76.20} \\
			\bottomrule
		\end{tabular}%
	}
	
\end{table*}

\begin{table}[t!]
	\centering
	\caption{Ablation study of the placement of
		AWS. † denotes the AWS before MHA and FF. * means the
		AWS between MHA and FF. ‡ is the AWS after MHA and FF.\label{posaws}}
	\setlength{\tabcolsep}{1.5mm}{
		\scriptsize
		\begin{tabular}{ccccrr}\toprule\multicolumn{2}{c}{\multirow{3}[6]{*}{Methods}} & \multicolumn{2}{c}{Attribution} & \multicolumn{2}{c}{Detection} \\\cmidrule(lr){3-4} \cmidrule(lr){5-6}\multicolumn{2}{c}{} & \multicolumn{2}{c}{Unseen} & \multicolumn{2}{c}{Unseen} \\\cmidrule(lr){3-4} \cmidrule(lr){5-6}\multicolumn{2}{c}{} & \multicolumn{2}{c}{ACC} & \multicolumn{1}{c}{ACC} & \multicolumn{1}{c}{AUC}\\\midrule\multicolumn{2}{l}{GazeCLIP w/AWS† } & \multicolumn{2}{c}{\textbf{80.66}} &  \textbf{ 89.80}   & 77.35  \\\multicolumn{2}{l}{GazeCLIP w/AWS*} & \multicolumn{2}{c}{71.66} &  89.68    & \textbf{77.53} \\\multicolumn{2}{l}{GazeCLIP w/AWS‡} & \multicolumn{2}{c}{79.34} &  89.25   & 70.24 \\\bottomrule\end{tabular}%
	}
	
\end{table}

\begin{table}[t!]
	\centering
	\caption{Effects of various gaze estimators. \label{tgaze2}}	
\setlength{\tabcolsep}{1.8mm}{
	\scriptsize
	\begin{tabular}{ccccrr}\toprule\multicolumn{2}{c}{\multirow{3}[6]{*}{Gaze Estimator}} & \multicolumn{2}{c}{Attribution} & \multicolumn{2}{c}{Detection} \\\cmidrule(lr){3-4} \cmidrule(lr){5-6}\multicolumn{2}{c}{} & \multicolumn{2}{c}{Unseen} & \multicolumn{2}{c}{Unseen} \\\cmidrule(lr){3-4} \cmidrule(lr){5-6}\multicolumn{2}{c}{} & \multicolumn{2}{c}{ACC} & \multicolumn{1}{c}{ACC} & \multicolumn{1}{c}{AUC}\\
			\midrule\multicolumn{2}{l}{Ours w/o gaze} & \multicolumn{2}{c}{65.87} &    80.97  & 63.67 \\
	\multicolumn{2}{l}{Park et al. \cite{Park}} & \multicolumn{2}{c}{68.20} &    76.13   & 70.57 \\
		
		\multicolumn{2}{l}{PureGaze \cite{PG}} & \multicolumn{2}{c}{72.13} &  80.44     & 75.89 \\\multicolumn{2}{l}{ETH-Xgaze \cite{ETHXGaze}} & \multicolumn{2}{c}{\textbf{80.66}} &  \textbf{ 89.80}   & \textbf{77.35} \\\bottomrule\end{tabular}%
}

\end{table}

\begin{table}[t!]
	\centering
\caption{Model ablation. \label{var}}
\setlength{\tabcolsep}{1.5mm}{
	\scriptsize
	\begin{tabular}{ccccccrr}\toprule\multicolumn{4}{c}{\multirow{2}[4]{*}{Model}} & \multicolumn{2}{c}{Attribution} & \multicolumn{2}{c}{Detection} \\\cmidrule(lr){5-6} \cmidrule(lr){7-8}\multicolumn{4}{c}{}          & \multicolumn{2}{c}{Unseen} & \multicolumn{2}{c}{Unseen} \\\cmidrule(lr){1-6} \cmidrule(lr){7-8} AGPM    & GE    & GIE   & LRE   & \multicolumn{2}{c}{ACC} & \multicolumn{1}{c}{ACC} & \multicolumn{1}{c}{AUC} \\\midrule  & checkmark
		&
		& 
		& \multicolumn{2}{c}{-}
		& 89.74
		& 40.83 \\ \checkmark    &       &       &       & \multicolumn{2}{c}{51.96} &   81.79    & 63.72 \\     \checkmark    &   \checkmark       &       &       & \multicolumn{2}{c}{62.03} &  83.82     & 65.41 \\        &      \checkmark   &  \checkmark       &   \checkmark      & \multicolumn{2}{c}{64.65} &   84.77    & 65.53 \\   \checkmark   &    \checkmark      &     \checkmark     &       & \multicolumn{2}{c}{67.99} &   86.78    & 68.26 \\    \checkmark     &     \checkmark     &       &    \checkmark      & \multicolumn{2}{c}{67.81} & 85.13      & 67.30 \\       \checkmark     &     \checkmark     &     \checkmark     &   \checkmark       & \multicolumn{2}{c}{\textbf{80.66}} &  \textbf{89.80}    & \textbf{77.35} \\   \bottomrule\end{tabular}%
}	
	
\end{table}

\begin{table}[t]
		\centering
			\renewcommand{\arraystretch}{1.2}
		\caption{Comparison of performance and computational efficiency on unseen datasets, including Celeb-DF \cite{Celeb-DF}, DFDC \cite{DFDC}, and WildDeepfake  \cite{wilddeepfake}, after training using seen generators and CelebA-HQ. \label{wild}}

		\setlength{\tabcolsep}{0.2mm}{
			\scriptsize
			\begin{tabular}{rrrrrrrrrrrrrr}\toprule\multicolumn{1}{c}{\multirow{3}[6]{*}{Method}} & \multicolumn{8}{c}{Attribution}                               & \multicolumn{2}{c}{Detection} & \multicolumn{1}{c}{\multirow{3}[6]{*}{\shortstack{Params \\(M)}}} & \multicolumn{1}{c}{\multirow{3}[6]{*}{\shortstack{FLOPs\\(G)} }} & \multicolumn{1}{c}{\multirow{3}[6]{*}{\shortstack{Inference\\Time (s)} }} \\ \cmidrule(lr){2-9} \cmidrule(lr){10-11}   & \multicolumn{2}{c}{Celeb-DF } & \multicolumn{2}{c}{DFDC } & \multicolumn{2}{c}{\shortstack{Wild\\Deepfake}} & \multicolumn{2}{c}{Average} & \multicolumn{2}{c}{Average} &       &       &  \\\cmidrule(lr){2-9} \cmidrule(lr){10-11}    & \multicolumn{2}{c}{ACC} & \multicolumn{2}{c}{ACC} & \multicolumn{2}{c}{ACC} & \multicolumn{2}{c}{ACC} & \multicolumn{1}{l}{ACC} & \multicolumn{1}{l}{AUC} &       &       &  \\\midrule\multicolumn{1}{c}{ResNet \cite{deep}} & \multicolumn{2}{c}{53.36} & \multicolumn{2}{c}{55.44} & \multicolumn{2}{c}{72.56} & \multicolumn{2}{c}{60.45} &  74.80     &   50.07    & \multicolumn{1}{c}{11.19} & \multicolumn{1}{c}{1.82} & \multicolumn{1}{c}{0.004} \\\multicolumn{1}{c}{Xception \cite{xception}} & \multicolumn{2}{c}{52.72} & \multicolumn{2}{c}{\underline{80.40}} & \multicolumn{2}{c}{\underline{81.92}} & \multicolumn{2}{c}{\underline{71.68}} &  \underline{75.06}     &    22.38   & \multicolumn{1}{c}{20.83} & \multicolumn{1}{c}{4.64} & \multicolumn{1}{c}{0.006} \\\multicolumn{1}{c}{ViT \cite{ViT}} & \multicolumn{2}{c}{62.88} & \multicolumn{2}{c}{70.88} & \multicolumn{2}{c}{63.76} & \multicolumn{2}{c}{65.84} &   \textbf{77.12}    &   \underline{74.86}   & \multicolumn{1}{c}{85.63 } & \multicolumn{1}{c}{16.88} & \multicolumn{1}{c}{0.007} \\\multicolumn{1}{c}{CViT \cite{CViT}} & \multicolumn{2}{c}{52.16} & \multicolumn{2}{c}{60.08} & \multicolumn{2}{c}{73.92} & \multicolumn{2}{c}{62.05} &    73.82   &    38.97   & \multicolumn{1}{c}{91.11} & \multicolumn{1}{c}{6.69} & \multicolumn{1}{c}{0.007} \\\multicolumn{1}{c}{CAEL \cite{genface}} & \multicolumn{2}{c}{44.08} & \multicolumn{2}{c}{31.12} & \multicolumn{2}{c}{39.52} & \multicolumn{2}{c}{38.24} &  71.78     &  53.19     & \multicolumn{1}{c}{158.65} & \multicolumn{1}{c}{2.11} & \multicolumn{1}{c}{0.111} \\\multicolumn{1}{c}{CLIP \cite{clip}} & \multicolumn{2}{c}{28.65} & \multicolumn{2}{c}{51.73} & \multicolumn{2}{c}{75.49} & \multicolumn{2}{c}{51.96} &     73.88  &  70.42     &\multicolumn{1}{c}{84.26}        &\multicolumn{1}{c}{15.91}  & \multicolumn{1}{c}{0.023} \\\multicolumn{1}{c}{MFCLIP \cite{MFCLIP}} & \multicolumn{2}{c}{\underline{78.08}} & \multicolumn{2}{c}{30.56} & \multicolumn{2}{c}{73.44} & \multicolumn{2}{c}{60.69} &  53.08     &    42.39   & \multicolumn{1}{c}{68.38 } & \multicolumn{1}{c}{7.17 } & \multicolumn{1}{c}{0.070} \\\multicolumn{1}{c}{DNA-Det \cite{DNADet}} & \multicolumn{2}{c}{18.72} & \multicolumn{2}{c}{47.60} & \multicolumn{2}{c}{4.24} & \multicolumn{2}{c}{23.52} &   75.00    &   23.20    & \multicolumn{1}{c}{5.90} & \multicolumn{1}{c}{16.37 } & \multicolumn{1}{c}{0.006}\\\multicolumn{1}{c}{DE-FAKE \cite{DEFAKE}} & \multicolumn{2}{c}{14.10} & \multicolumn{2}{c}{25.38} & \multicolumn{2}{c}{28.76} & \multicolumn{2}{c}{22.75} &   70.31    &  35.69   & \multicolumn{1}{c}{84.26} & \multicolumn{1}{c}{95.91 } & \multicolumn{1}{c}{0.053} \\\multicolumn{1}{c}{FA \cite{ForensicsAdapter}} & \multicolumn{2}{c}{27.60} & \multicolumn{2}{c}{47.52 } & \multicolumn{2}{c}{56.08 } & \multicolumn{2}{c}{43.73} &    73.18   &  44.07     &    \multicolumn{1}{c}{5.70}   & \multicolumn{1}{c}{-} & \multicolumn{1}{c}{0.049} \\ \multicolumn{1}{c}{Lin et al. \cite{Lin}} & \multicolumn{2}{c}{30.56} & \multicolumn{2}{c}{50.79} & \multicolumn{2}{c}{72.80} & \multicolumn{2}{c}{ 51.38} &   73.93    &   50.84    & \multicolumn{1}{c}{85.31} & \multicolumn{1}{c}{96.04 } & \multicolumn{1}{c}{0.057}
				\\\multicolumn{1}{c}{OmniDFA \cite{omnidfa}} & \multicolumn{2}{c}{45.76} & \multicolumn{2}{c}{48.09} & \multicolumn{2}{c}{50.11} & \multicolumn{2}{c}{47.99} &   74.17    &   48.95    & \multicolumn{1}{c}{31.75} & \multicolumn{1}{c}{5.88 } & \multicolumn{1}{c}{0.006}\\\multicolumn{1}{c}{CDAL \cite{cdal}} & \multicolumn{2}{c}{33.79} & \multicolumn{2}{c}{40.66} & \multicolumn{2}{c}{43.80} & \multicolumn{2}{c}{39.42} &   72.04    &   40.61    & \multicolumn{1}{c}{38.26} & \multicolumn{1}{c}{1.04 } & \multicolumn{1}{c}{0.006}\\ \multicolumn{1}{c}{GazeCLIP(Ours)} & \multicolumn{2}{c}{\textbf{93.52}} & \multicolumn{2}{c}{\textbf{94.88}} & \multicolumn{2}{c}{\textbf{99.68}} & \multicolumn{2}{c}{\textbf{96.03}} &    74.88   &   \textbf{94.20}    &   \multicolumn{1}{c}{204.40}  &     \multicolumn{1}{c}{17.65} &  \multicolumn{1}{c}{0.192} \\
				\midrule   
			\end{tabular}%
		}

	\hfill
		\centering
		\caption{ Comparison of performance on unseen advanced generators, including VAE \cite{vae}, HART \cite{hart}, and FLUX \cite{flux}, after training using seen generators and CelebA-HQ. \label{flux}}
		\setlength{\tabcolsep}{2.0mm}{
			\scriptsize
			\begin{tabular}{rrrrrrrrrrr}      &       &       &       &       &       &       &       &       &       &  \\\midrule\multicolumn{1}{c}{\multirow{3}[6]{*}{Method}} & \multicolumn{8}{c}{ Attribution}                              & \multicolumn{2}{c}{Detection} \\ \cmidrule(lr){2-9} \cmidrule(lr){10-11}  & \multicolumn{2}{c}{VAE } & \multicolumn{2}{c}{HART } & \multicolumn{2}{c}{FLUX } & \multicolumn{2}{c}{Average} & \multicolumn{2}{c}{Average} \\\cmidrule(lr){2-9}   \cmidrule(lr){10-11}  & \multicolumn{2}{c}{ACC} & \multicolumn{2}{c}{ACC} & \multicolumn{2}{c}{ACC} & \multicolumn{2}{c}{ACC} & \multicolumn{1}{c}{ACC} & \multicolumn{1}{c}{AUC} \\ \cmidrule(lr){2-11}  \multicolumn{1}{c}{ResNet \cite{deep} } & \multicolumn{2}{c}{58.56} & \multicolumn{2}{c}{\underline{78.64}} & \multicolumn{2}{c}{\underline{52.96}} & \multicolumn{2}{c}{\underline{63.39}} &   \underline{51.08}    & 35.30 \\\multicolumn{1}{c}{Xception \cite{xception}} & \multicolumn{2}{c}{\textbf{84.16}} & \multicolumn{2}{c}{20.24} & \multicolumn{2}{c}{19.20} & \multicolumn{2}{c}{41.20} &    \underline{51.08}   & 26.67 \\\multicolumn{1}{c}{ViT \cite{ViT}} & \multicolumn{2}{c}{53.52} & \multicolumn{2}{c}{14.08} & \multicolumn{2}{c}{15.44} & \multicolumn{2}{c}{27.68} & 42.76      &39.72  \\\multicolumn{1}{c}{CViT \cite{CViT}} & \multicolumn{2}{c}{3.36} & \multicolumn{2}{c}{26.00} & \multicolumn{2}{c}{15.84} & \multicolumn{2}{c}{15.07} &   42.90    & 28.68 \\\multicolumn{1}{c}{CLIP \cite{clip}} & \multicolumn{2}{c}{59.48} & \multicolumn{2}{c}{35.72} & \multicolumn{2}{c}{30.69} & \multicolumn{2}{c}{41.96} &   40.63    &38.79  \\\multicolumn{1}{c}{CAEL \cite{genface}} & \multicolumn{2}{c}{34.48} & \multicolumn{2}{c}{29.76} & \multicolumn{2}{c}{31.28 } & \multicolumn{2}{c}{31.84 } &   33.16    & 17.68 \\\multicolumn{1}{c}{MFCLIP \cite{MFCLIP}} & \multicolumn{2}{c}{6.96} & \multicolumn{2}{c}{20.56} & \multicolumn{2}{c}{14.72 } & \multicolumn{2}{c}{14.08} &  37.94     & 36.67 \\\multicolumn{1}{c}{DNA-Det \cite{DNADet}} & \multicolumn{2}{c}{\underline{80.16}} & \multicolumn{2}{c}{39.44} & \multicolumn{2}{c}{17.60} & \multicolumn{2}{c}{45.73} &  44.82     & 15.58\\\multicolumn{1}{c}{DE-FAKE \cite{DEFAKE}} & \multicolumn{2}{c}{4.96} & \multicolumn{2}{c}{12.58} & \multicolumn{2}{c}{3.75} & \multicolumn{2}{c}{7.10} &  20.89     & 13.75 \\\multicolumn{1}{c}{FA \cite{ForensicsAdapter}} & \multicolumn{2}{c}{59.92} & \multicolumn{2}{c}{54.72} & \multicolumn{2}{c}{38.32} & \multicolumn{2}{c}{50.99} &  40.37     & 39.24
				\\\multicolumn{1}{c}{Lin et al. \cite{Lin}} & \multicolumn{2}{c}{60.34} & \multicolumn{2}{c}{37.41} & \multicolumn{2}{c}{40.66} & \multicolumn{2}{c}{46.14} &  44.67     & \underline{40.30} \\		
				\multicolumn{1}{c}{OmniDFA \cite{omnidfa}} & \multicolumn{2}{c}{40.71} & \multicolumn{2}{c}{38.62} & \multicolumn{2}{c}{30.29} & \multicolumn{2}{c}{36.54} &  40.67     & 39.43    \\   \multicolumn{1}{c}{CDAL \cite{cdal}} & \multicolumn{2}{c}{37.89} & \multicolumn{2}{c}{35.61} & \multicolumn{2}{c}{20.83} & \multicolumn{2}{c}{31.44} &  35.20     & 30.96 \\ GazeCLIP (Ours)     &\multicolumn{2}{c}{75.44} & \multicolumn{2}{c}{\textbf{84.88}} & \multicolumn{2}{c}{\textbf{71.44}} & \multicolumn{2}{c}{\textbf{77.25}} & \textbf{51.18}      & \textbf{54.45} \\  \midrule      
			\end{tabular}%
		}
\end{table}

\begin{table}[t!]
	\centering 
	\caption{Ablations of the plug-and-play AWS.  \label{aws}} 
	\setlength{\tabcolsep}{1.5mm}{
		\scriptsize
		\begin{tabular}{ccccrrrr}\toprule\multicolumn{2}{c}{\multirow{3}[6]{*}{Model}} & \multicolumn{2}{c}{Attribution} & \multicolumn{2}{c}{Detection} & \multicolumn{2}{c}{\multirow{2}[4]{*}{}} \\\cmidrule(lr){3-4} \cmidrule(lr){5-6}\multicolumn{2}{c}{} & \multicolumn{2}{c}{Unseen} & \multicolumn{2}{c}{Unseen} & \multicolumn{2}{c}{} \\\cmidrule(lr){3-4} \cmidrule(lr){5-8}\multicolumn{2}{c}{} & \multicolumn{2}{c}{ACC} & \multicolumn{1}{c}{ACC} & \multicolumn{1}{c}{AUC} & \multicolumn{1}{c}{\shortstack{Params\\(M)} } & \multicolumn{1}{c}{\shortstack{FLOPs\\(G)} } \\\midrule\multicolumn{2}{c}{CLIP w/o AWS} & \multicolumn{2}{c}{74.10} &    65.55   &   69.91    &  84.26     & \multicolumn{1}{c}{15.91}  \\\multicolumn{2}{c}{CLIP w/ AWS} & \multicolumn{2}{c}{78.36} &     73.29 &    70.26   & 84.26      & \multicolumn{1}{c}{15.91}   \\\midrule\multicolumn{2}{c}{GazeCLIP (Ours) w/o AWS} &  \multicolumn{2}{c}{60.42} &     \multicolumn{1}{c}{87.68}    &   \multicolumn{1}{c}{ 68.31} &  204.40     & \multicolumn{1}{c}{17.65}  \\\multicolumn{2}{c}{GazeCLIP (Ours) w/ AWS} & \multicolumn{2}{c}{\textbf{80.66}} &     \multicolumn{1}{c}{\textbf{89.80}}    &   \multicolumn{1}{c}{\textbf{77.35}}   & 204.40     &  \multicolumn{1}{c}{17.65}\\\bottomrule\end{tabular}%
	}	
\end{table}

\begin{figure*}[t!]
	\centering
	\includegraphics[width=\linewidth]{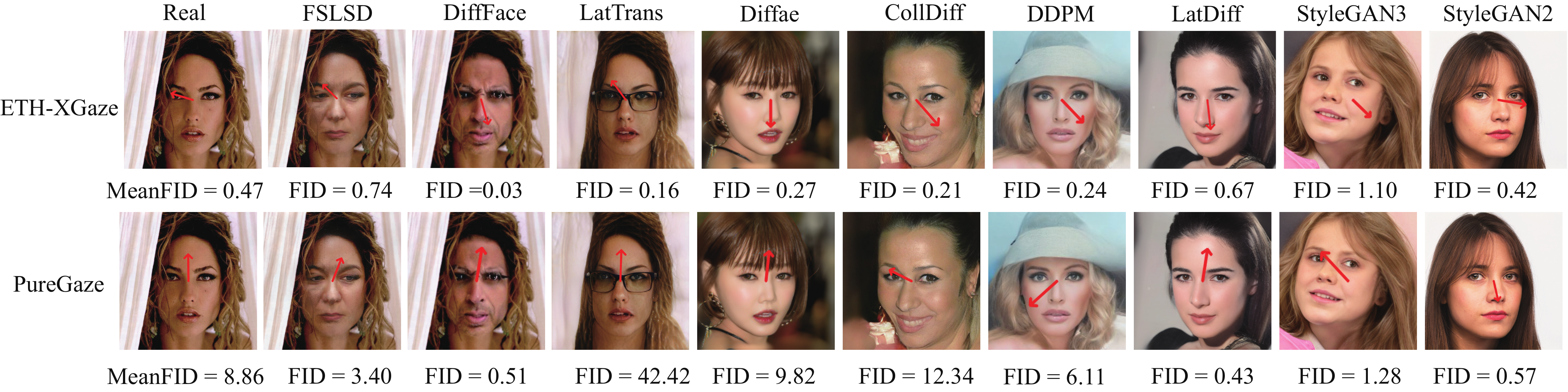}
	\caption{Visualization of gaze embeddings produced by various gaze estimators, including ETH-XGaze and PureGaze, for fake images synthesized by various generators. We randomly select 10k face images and 10k fake ones to extract gaze embeddings using various pre-trained gaze estimators, and calculate the FID score between real and fake gaze embeddings. Lower FID indicates a smaller distribution gap between real and fake gaze embeddings.}\label{oga}
\end{figure*}

{\bfseries\setlength\parindent{0em} Benchmark results.} We test models on various unseen generators and FFHQ after training using both seen generators and CelebA-HQ. In Table~\ref{bench}, the ACC of our method exceeds that of most SOTAs, showing the excellent attribution and detection capability of our method. Specifically, the attribution average ACC score of our GazeCLIP model is about 6\% higher than that of CLIP on unseen generators. Unlike CLIP, which only mines global image and language forgery embeddings, our network focuses on general gaze-aware image forgery features and performs fine-grained vision-language alignment. Besides, the ACC of model detection generally exceeds that of model attribution. For instance, CAEL and DNA-Det achieve the average ACC score of 25.74\% and 33.14\%, accordingly. By contrast, they attain 81.71\% and 86.99\% ACC for detection. Thus, attribution tends to be more difficult than detection. We further test models on various more recent advanced generators and FFHQ after training using both seen generators and CelebA-HQ. In Table~\ref{bench2}, the attribution average ACC score of our GazeCLIP model is about 20\% higher than that of the lightweight Xception. Different from traditional CNN methods that focus on low-level texture features with limited generalizability, our model pays attention to the high-level gaze semantic features which effectively reflect the generator's ability to model facial semantics, and are more suitable for tracing the source of deepfakes. CLIP achieves relatively strong performance on both tasks, with an attribution ACC of 70.69\% and a detection AUC of 71.64\%, while DE-FAKE, OmniDFA, and CDAL maintain moderate detection performance but show limited attribution capability. In contrast, GazeCLIP achieves consistently strong performance on both tasks, reaching 86.41\% attribution ACC and 81.23\% detection AUC. This indicates that GazeCLIP learns more transferable representations that capture both general forgery cues for detection and generator-specific characteristics for attribution.

Following the protocol described in Table~\ref{bench}, FFHQ contains pristine real face images without forgery-attribution labels, and we use it as an unseen real-domain stress test to examine whether the model exhibits spurious generator-attribution behaviour. Therefore, the low attribution accuracy on FFHQ (5.12\%) does not mean poor generalization of GazeCLIP. Since FFHQ contains pristine real images without a valid generator identity in the attribution label space, this result is not directly comparable to attribution accuracy on forged samples and should not be used alone to assess the effectiveness of the gaze prior. Similar behaviour can also be observed in other attribution models, where the attribution performance on FFHQ remains low due to the domain gap between real and forged images. Moreover, the effectiveness of the gaze prior is mainly reflected in distinguishing generator-specific forgery patterns on unseen generators rather than assigning labels to pristine real faces. To further examine the generalization of GazeCLIP to high-quality real face datasets and the effectiveness of the gaze prior in such scenarios, we evaluate the real/fake detection performance of GazeCLIP with and without gaze on FFHQ. GazeCLIP with gaze achieves an AUC of 85.12\%, compared with 68.22\% for GazeCLIP without gaze, demonstrating that the gaze prior remains beneficial under the unseen high-quality real-image domain.

{\bfseries\setlength\parindent{0em} Generalization to unseen in-the-wild datasets.}
To provide stronger evidence of domain generalization beyond synthetic benchmarks, we train networks using our training sets and test them on unseen in-the-wild datasets, including WildDeepfake \cite{wilddeepfake}, Celeb-DF \cite{Celeb-DF}, and DFDC \cite{DFDC}. In Table~\ref{wild}, both attribution and detection performance of our method outperform that of most existing methods, showing the strong capability in domain generalization. To conduct the runtime benchmark, we evaluate the computational cost, including the parameter count, FLOPs, and inference time. Despite more parameters than DNA-Det, GazeCLIP shows significant improvements in both attribution and detection performance. In particular, the attribution average ACC of GazeCLIP is about 72.51\% higher than that of lightweight DNA-Det, with an increase of 198.50M parameters and 1.28G FLOPs. In the future, we plan to leverage knowledge distillation or model adjustments to enhance efficiency. We also test models on face images synthesized by HART, VAE, and flow-based FLUX models. In Table~\ref{flux},  methods tend to obtain the lowest ACC score on face images generated by FLUX, among all generators. This indicates that FLUX generates high-quality face images that may be more complex and challenging.

{\bfseries\setlength\parindent{0em} Forgery type traceability}.
We further explore the forgery type traceability of models on our benchmark. Specifically, we formulate forgery-type traceability as a four-class classification task, with Real, EFS, FS, and AM assigned labels 0, 1, 2, and 3, respectively. For various state-of-the-art methods, we introduce a forgery-type trace expert as an additional branch alongside the detection and attribution experts. We train all methods using seen generators together with real images from CelebA-HQ, and evaluate their ability to identify forgery types on unseen generators from DF40, with FFHQ serving as the unseen real-image source. As shown in Table~\ref{ftbench}, the attribution average ACC of GazeCLIP is about 13.6\% higher than that of CLIP. In particular, GazeCLIP achieves consistently high forgery type attribution accuracy across most unseen generators, reaching over 90\% ACC on SimSwap, InSwapper, and SD-2.1. This demonstrates that our benchmark can effectively support forgery-type traceability evaluation and further reveal the ability of forensic models to capture forgery-type-related cues on unseen generators.

{\bfseries\setlength\parindent{0em} Robustness analysis.} 
We train networks using our protocol and test their attribution performance on unseen distorted images from \cite{df10}. In Figure~\ref{robust2}, ten types of corruption are engaged, each with five severities. A severity of 0 means no corruption. In detail, specific settings for the intensity of ten deformations are discussed in \cite{genface}. Corruption types and severities are applied only during testing and are not used during training. Models are inclined to show sensitivity to deformations like video compression. Our approach significantly outperforms most models across various types of image perturbations. Different from existing methods that focus on low-level texture features that are highly susceptible to image quality degradation, our model considers high-level gaze features related to the consistency of facial pose, eye movement, and facial expression, which makes our method equip strong robustness to image quality variations. 
\subsection{Ablation Study}
{\bfseries\setlength\parindent{0em} Impacts of various modules.}
In Table~\ref{var}, GE increases the attribution performance by about 10\% ACC, showing that gaze features offer valuable clues to boost DFAD. AGPM enhances the ACC by about 16\%, which validates that appearance-gaze global manipulated embeddings are vital for DFAD. The gain from introducing GIE (+5.96\%) is obvious, emphasizing the importance of general gaze-aware image forgery embeddings. LRE improves performance by about 18.63\% ACC when GIE is also involved. The gain brought by LRE alone (+5.78\%) is marginal without the introduction of GIE, as the effectiveness of LRE depends on the synergy between the various common forgery features. LRE offers fine-grained adaptive-enhanced semantic guidance, which is enhanced when combined with the distinctive gaze prior and general facial image forgery features from GIE. Without the module, LRE's potential is not fully realized, leading to a smaller performance gain. These results show that language guidance is more effective when gaze and general image forgery features are jointly incorporated, as their complementary visual cues provide a richer foundation for language-guided fine-grained forgery representation learning.

{\bfseries\setlength\parindent{0em} Effects of gaze estimator.}
We delve into the effect of various gaze estimators, including ETH-Xgaze \cite{ETHXGaze}, PureGaze \cite{PG}, and Park et al. \cite{Park}. Specifically, we visualize the gaze embeddings produced by different gaze estimators and compute the FID between 10k gaze embeddings extracted from real and fake images synthesized by different generators. A lower FID indicates a smaller distributional discrepancy between real and fake gaze embeddings. As shown in Fig.~\ref{oga}, different gaze estimators exhibit distinct distributional characteristics. Although different gaze estimators may produce various predictions for the same fake face image, we do not regard their outputs as accurate gaze ground truth. Instead, we treat them as high-level gaze-sensitive representations, or auxiliary forensic priors, extracted by pretrained gaze estimators. Therefore, our objective is not to recover the exact gaze state of a forged face, but to investigate whether the extracted gaze-sensitive representations capture systematic
differences between real and forged faces. In this context, the FID provides a distribution-level measure of the discrepancy between real and fake gaze representations, which is less dependent on the prediction of any individual sample, and allows us to characterize how the distribution of gaze-sensitive representations varies across different estimators. 

To further examine the robustness of the discovered distribution patterns, we do not rely solely on the averaged FID or the output of a single gaze estimator. Instead, we independently evaluate different gaze estimators and compare their attribution and detection performance. As shown in Table~\ref{tgaze2}, the model without gaze achieves 65.87\% attribution ACC and 68.22\% detection AUC on unseen generators, while incorporating different pretrained gaze estimators consistently improves both attribution and detection performance. This suggests that the effectiveness of
the gaze prior does not solely rely on the absolute accuracy of individual gaze predictions. Instead, gaze-sensitive representations can provide complementary forensic information for characterizing generator-related artifacts. Furthermore, although ETH-XGaze yields a lower FID than PureGaze, it achieves the highest attribution ACC and detection AUC, indicating that forensic performance depends not on the magnitude of distributional discrepancy but on the stability and discriminability of forgery-related cues. Overall, the consistent improvements across different pretrained gaze estimators provide additional evidence that GazeCLIP benefits from robust gaze-sensitive forensic features rather than simply exploiting estimator-specific errors or distribution shifts.


{\bfseries\setlength\parindent{0em} Impacts of adaptive-enhanced word selector.}
To evaluate the efficiency of AWS, we conduct ablations to show that AWS can generalize to various transformer-based vision-language models for performance improvement. In Table~\ref{aws}, due to the addition of AWS, the attribution ACC of both CLIP and GazeCLIP is increased by 4.26\% and 20.24\%, respectively. Besides, AWS rarely adds auxiliary weights and computational costs when added to other models. We visualize the language feature matrix of CLIP and GazeCLIP, when AWS is involved or not. For CLIP, we only introduce the attribution prompt. In Figure~\ref{dwa}, important word embeddings are emphasized while unimportant ones are suppressed, due to the addition of AWS, showing that AWS achieves flexible language refinement. Compared to other prompts, the last prompt is highly reinforced, indicating its value for DFAD.

\begin{figure*}[t]
	\centering
	\includegraphics[width=\linewidth]{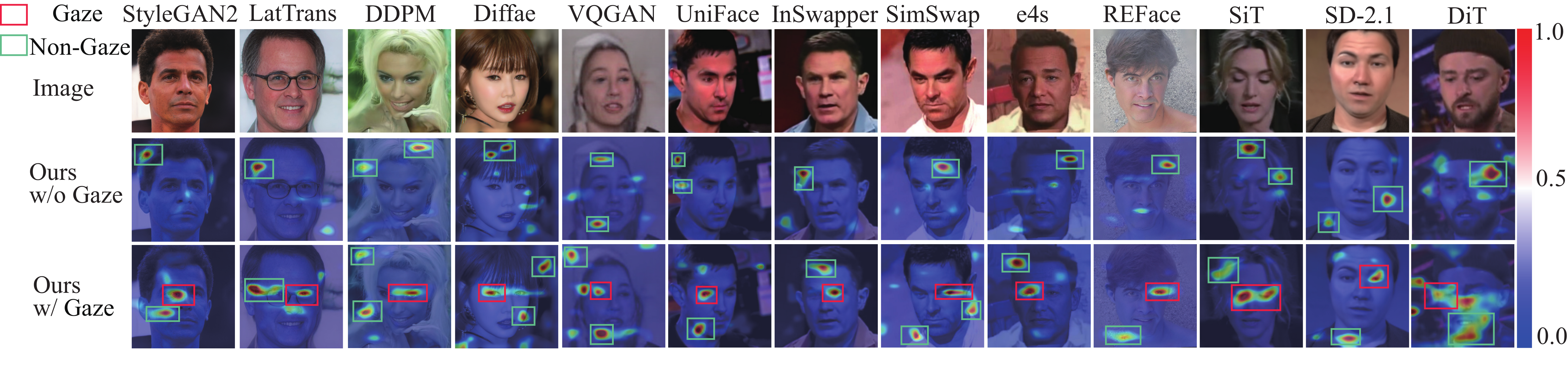}
	\caption{The heatmap visualizations of our GazeCLIP model (w/o or w/ gaze) on different generator samples. The hotter (red color) a position is, the more forgery features are captured by the network.}
	\label{Fig4heatmap}
\end{figure*}

{\bfseries\setlength\parindent{0em} Influences of fine-grained text prompts.} 
To study the impact of fine-grained text prompts, we evaluate the performance of GazeCLIP by gradually adding hierarchical texts and learnable language prompts with insufficient accuracy. In Table~\ref{lp}, we note that the performance of our model tends to improve with the growth of hierarchical texts. The attribution ACC of GazeCLIP declines (-5.3\%) due to the addition of learnable text prompts. Therefore, detailed and precise semantic guidance is crucial for the fine-grained DFAD task.

\begin{figure}[t]
	\centering
	\includegraphics[width=\linewidth]{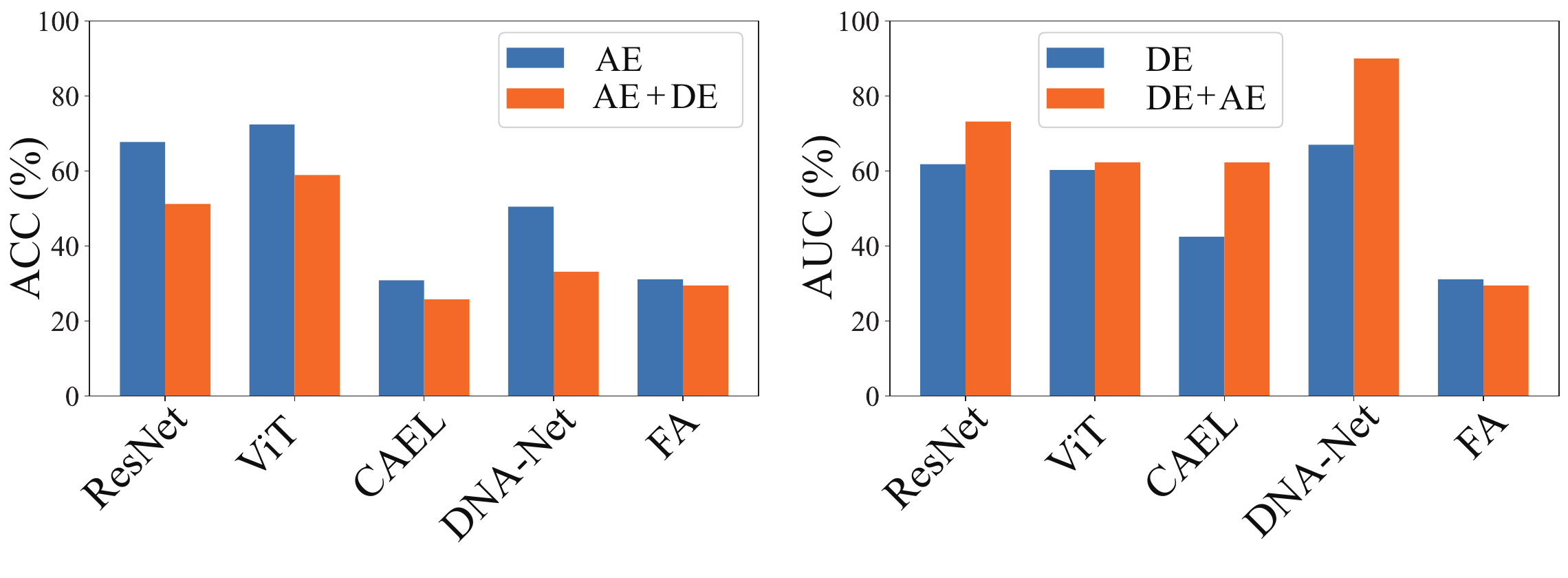}
	\caption{Left: The attribution performance of various DFAD models (only w/ AE or w/ AE+DE). Right: The detection performance of various DFAD models (only w/ DE or w/ AE+DE). AE and DE are attribution expert and detection expert, respectively. }
	\label{dfadfd}
\end{figure}

\begin{table}[t!]
	\centering
	\caption{Ablations of various queries. Results are averaged over unseen generators for attribution and detection. \label{query}}
	\setlength{\tabcolsep}{1.0mm}{
		\small
		\begin{tabular}{ccccrrrr}\toprule\multicolumn{2}{c}{\multirow{3}[6]{*}{Model}} & \multicolumn{2}{c}{Attribution} & \multicolumn{2}{c}{Detection} & \multirow{2}[4]{*}{} & \multirow{2}[4]{*}{} \\\cmidrule(lr){3-4} \cmidrule(lr){5-6}\multicolumn{2}{c}{} & \multicolumn{2}{c}{Unseen} & \multicolumn{2}{c}{Unseen} &       &  \\\cmidrule{3-8}\multicolumn{2}{c}{} & \multicolumn{2}{c}{ACC} & \multicolumn{1}{c}{ACC} & \multicolumn{1}{c}{AUC} & \multicolumn{1}{l}{\shortstack{Params\\ (M)}} & \multicolumn{1}{l}{\shortstack{FLOPs\\ (G)}} \\\midrule  \multicolumn{2}{c}{Query w/patch} & \multicolumn{2}{c}{78.96} & 88.94      &  76.73     & 204.40      & 18.54 \\\multicolumn{2}{c}{Query w/all} & \multicolumn{2}{c}{\textbf{81.23}} &   88.88    &   72.48    & 204.40      &18.56 \\\multicolumn{2}{c}{Query w/cls} & \multicolumn{2}{c}{80.66} &    \textbf{89.80}   &    \textbf{77.35}  &   204.40    & 17.50\\\bottomrule\end{tabular}%
	}
	
\end{table}

\begin{table}[t!]
	\centering
	\caption{Ablations of various language prompts. LP is the learnable language prompts. Results are averaged over unseen generators for attribution and detection. \label{lp}}
	\setlength{\tabcolsep}{1.9mm}{
		\small
		\begin{tabular}{ccccrr}\toprule\multicolumn{2}{c}{\multirow{3}[6]{*}{Model}} & \multicolumn{2}{c}{Attribution} & \multicolumn{2}{c}{Detection} \\\cmidrule(lr){3-4} \cmidrule(lr){5-6}\multicolumn{2}{c}{} & \multicolumn{2}{c}{Unseen} & \multicolumn{2}{c}{Unseen} \\\cmidrule(lr){3-4} \cmidrule(lr){5-6}\multicolumn{2}{c}{} & \multicolumn{2}{c}{ACC} & \multicolumn{1}{c}{ACC} & \multicolumn{1}{c}{AUC}\\\midrule\multicolumn{2}{l}{L1} &  \multicolumn{2}{c}{72.76} &    84.07   & 61.73 \\\multicolumn{2}{l}{L1+L2} & \multicolumn{2}{c}{78.03} &   85.38    & 63.59 \\\multicolumn{2}{l}{L1+L2+L3} & \multicolumn{2}{c}{67.80} &    86.26   & 65.25 \\\multicolumn{2}{l}{L1+L2+L3+L4} & \multicolumn{2}{c}{\textbf{80.66}} &   \textbf{89.80}    & \textbf{77.35} \\\multicolumn{2}{l}{L1+L2+L3+L4+LP} &  \multicolumn{2}{c}{75.37} &   87.64    &  70.32\\\bottomrule\end{tabular}%
	}
	
\end{table}

\begin{table*}[t!]
	\centering
	\caption{Attribution and detection performance on unseen generators under source-matched, identity-matched, head-pose-matched, and image-quality-degraded testing settings. Attr.: Attribution; Det.: Detection.}
	\label{tab:control_ablation}
	\setlength{\tabcolsep}{1.0mm}
	\small
	
	\begin{tabular}{lccccccccccccccc}
		\toprule
		
		\multirow{4}{*}{Model}
		& \multicolumn{3}{c}{Original}
		& \multicolumn{3}{c}{Source-matched}
		& \multicolumn{3}{c}{Identity-matched}
		& \multicolumn{3}{c}{Head-pose-matched}
		& \multicolumn{3}{c}{Image-quality-degraded} \\
		
		\cmidrule(lr){2-4}
		\cmidrule(lr){5-7}
		\cmidrule(lr){8-10}
		\cmidrule(lr){11-13}
		\cmidrule(lr){14-16}
		
		&
		\multicolumn{1}{c}{Attr.}
		& \multicolumn{2}{c}{Det.}
		&
		\multicolumn{1}{c}{Attr.}
		& \multicolumn{2}{c}{Det.}
		&
		\multicolumn{1}{c}{Attr.}
		& \multicolumn{2}{c}{Det.}
		&
		\multicolumn{1}{c}{Attr.}
		& \multicolumn{2}{c}{Det.}
		&
		\multicolumn{1}{c}{Attr.}
		& \multicolumn{2}{c}{Det.} \\
		
		\cmidrule(lr){2-2}
		\cmidrule(lr){3-4}
		\cmidrule(lr){5-5}
		\cmidrule(lr){6-7}
		\cmidrule(lr){8-8}
		\cmidrule(lr){9-10}
		\cmidrule(lr){11-11}
		\cmidrule(lr){12-13}
		\cmidrule(lr){14-14}
		\cmidrule(lr){15-16}
		
		&
		\multicolumn{1}{c}{Unseen}
		& \multicolumn{2}{c}{Unseen}
		&
		\multicolumn{1}{c}{Unseen}
		& \multicolumn{2}{c}{Unseen}
		&
		\multicolumn{1}{c}{Unseen}
		& \multicolumn{2}{c}{Unseen}
		&
		\multicolumn{1}{c}{Unseen}
		& \multicolumn{2}{c}{Unseen}
		&
		\multicolumn{1}{c}{Unseen}
		& \multicolumn{2}{c}{Unseen} \\
		
		\cmidrule(lr){2-2}
		\cmidrule(lr){3-4}
		\cmidrule(lr){5-5}
		\cmidrule(lr){6-7}
		\cmidrule(lr){8-8}
		\cmidrule(lr){9-10}
		\cmidrule(lr){11-11}
		\cmidrule(lr){12-13}
		\cmidrule(lr){14-14}
		\cmidrule(lr){15-16}
		
		&
		ACC & ACC & AUC
		& ACC & ACC & AUC
		& ACC & ACC & AUC
		& ACC & ACC & AUC
		& ACC & ACC & AUC \\
		
		\midrule
		
		Ours w/o Gaze
		& 65.87 & 80.97 & 68.22
		& 71.02 & 80.85 & 62.85
		& 70.69 & 79.17 & 60.48
		& 67.20 & 79.63 & 66.41
		& 63.90 & 78.45 & 64.92 \\
		
		GazeCLIP (Ours)
		& \textbf{80.66}
		& \textbf{89.80}
		& \textbf{77.35}
		& \textbf{86.46} & \textbf{89.76} & \textbf{74.00}
		& \textbf{86.10} & \textbf{88.89} & \textbf{73.90}
		& \textbf{79.10} & \textbf{87.82} & \textbf{75.26}
		& \textbf{80.57} & \textbf{89.78} & \textbf{77.14} \\
		
		\bottomrule
	\end{tabular}
\end{table*}

\begin{table*}[t!]
	\centering
	\caption{Attribution and detection performance on unseen generators under different region-masking testing settings. Attr.: Attribution; Det.: Detection.}
	\label{reginmask}
	\setlength{\tabcolsep}{1.0mm}
	\small
	
\begin{tabular}{lccccccccccccccc}
	\toprule
	
	\multirow{4}{*}{Model}
	& \multicolumn{3}{c}{Original}
	& \multicolumn{3}{c}{Eye-masked}
	& \multicolumn{3}{c}{Mouth-masked}
	& \multicolumn{3}{c}{Face-masked}
	& \multicolumn{3}{c}{Background-masked} \\
	
	\cmidrule(lr){2-4}
	\cmidrule(lr){5-7}
	\cmidrule(lr){8-10}
	\cmidrule(lr){11-13}
	\cmidrule(lr){14-16}
	
	&
	\multicolumn{1}{c}{Attr.}
	& \multicolumn{2}{c}{Det.}
	&
	\multicolumn{1}{c}{Attr.}
	& \multicolumn{2}{c}{Det.}
	&
	\multicolumn{1}{c}{Attr.}
	& \multicolumn{2}{c}{Det.}
	&
	\multicolumn{1}{c}{Attr.}
	& \multicolumn{2}{c}{Det.}
	&
	\multicolumn{1}{c}{Attr.}
	& \multicolumn{2}{c}{Det.} \\
	
	\cmidrule(lr){2-2}
	\cmidrule(lr){3-4}
	\cmidrule(lr){5-5}
	\cmidrule(lr){6-7}
	\cmidrule(lr){8-8}
	\cmidrule(lr){9-10}
	\cmidrule(lr){11-11}
	\cmidrule(lr){12-13}
	\cmidrule(lr){14-14}
	\cmidrule(lr){15-16}
	
	&
	\multicolumn{1}{c}{Unseen}
	& \multicolumn{2}{c}{Unseen}
	&
	\multicolumn{1}{c}{Unseen}
	& \multicolumn{2}{c}{Unseen}
	&
	\multicolumn{1}{c}{Unseen}
	& \multicolumn{2}{c}{Unseen}
	&
	\multicolumn{1}{c}{Unseen}
	& \multicolumn{2}{c}{Unseen}
	&
	\multicolumn{1}{c}{Unseen}
	& \multicolumn{2}{c}{Unseen} \\
	
	\cmidrule(lr){2-2}
	\cmidrule(lr){3-4}
	\cmidrule(lr){5-5}
	\cmidrule(lr){6-7}
	\cmidrule(lr){8-8}
	\cmidrule(lr){9-10}
	\cmidrule(lr){11-11}
	\cmidrule(lr){12-13}
	\cmidrule(lr){14-14}
	\cmidrule(lr){15-16}
	
	&
	ACC & ACC & AUC
	& ACC & ACC & AUC
	& ACC & ACC & AUC
	& ACC & ACC & AUC
	& ACC & ACC & AUC \\
	
	\midrule
	
	Ours w/o Gaze
	& 65.87 & 80.97 & 68.22
	& \textbf{71.71} & 81.23 & \textbf{75.71}
	& \textbf{75.97} & 81.18 & \textbf{74.00}
	& 77.64 & 81.09 & 69.87
	& \textbf{79.38} & 81.12 & 71.39 \\
	
	GazeCLIP (Ours)
	& \textbf{80.66}
	& \textbf{89.80}
	& \textbf{77.35}
	& 40.43 & \textbf{87.58} & 70.65
	& 60.75 & \textbf{89.65} & 73.82
	& \textbf{80.72} & \textbf{89.00} & \textbf{74.58}
	& 60.46 & \textbf{89.78} & \textbf{77.14} \\
	
	\bottomrule
\end{tabular}
\end{table*}

\begin{table}[t!]
	\centering
	\caption{Ablation study of different numbers of LoRAs. Results are averaged over unseen generators for attribution and detection. \label{lora}}
	\setlength{\tabcolsep}{1.9mm}{
		\small
		\begin{tabular}{ccccrr}\toprule\multicolumn{2}{c}{\multirow{2}[4]{*}{LoRA Number}} & \multicolumn{2}{c}{Attribution} & \multicolumn{2}{c}{Detection} \\\cmidrule(lr){3-4} \cmidrule(lr){5-6}\multicolumn{2}{c}{} & \multicolumn{2}{c}{Unseen} & \multicolumn{2}{c}{Unseen} \\\midrule GIE   & LRE   & \multicolumn{2}{c}{ACC} & \multicolumn{1}{c}{ACC} & \multicolumn{1}{c}{AUC} \\\midrule1     & 1     & \multicolumn{2}{c}{78.86} &    85.27   & 46.48 \\2     & 1     & \multicolumn{2}{c}{\textbf{80.66}} &   89.80    & \textbf{77.35} \\3     & 1     & \multicolumn{2}{c}{80.07} &   89.70    & 66.45 \\4     & 1     & \multicolumn{2}{c}{71.13} &  \textbf{90.00}    & 76.15 \\3     & 2     & \multicolumn{2}{c}{68.50} &   87.45    & 58.69 \\\bottomrule\end{tabular}%
	}

\end{table}
\begin{figure}[t!]
	\centering
	\includegraphics[width=\linewidth]{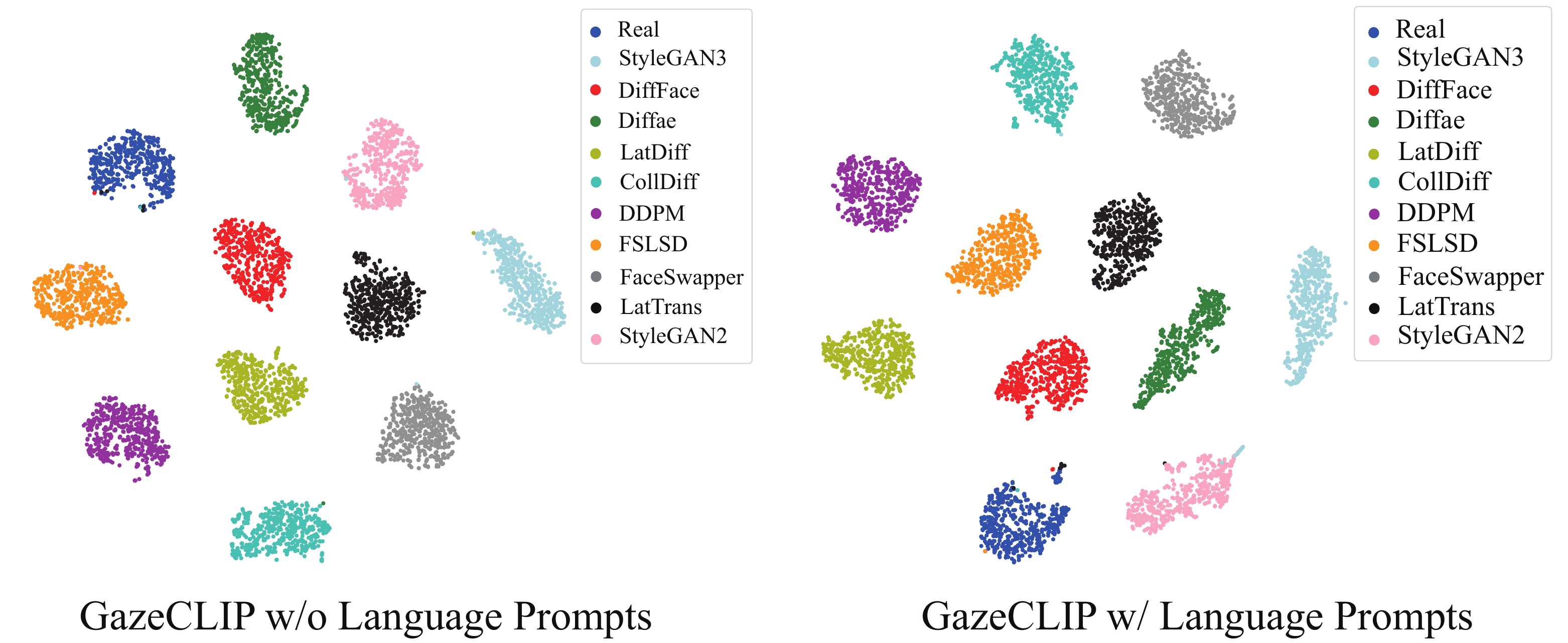} 
	\caption{The t-SNE visualization of various generator distributions generated by GazeCLIP without or with language prompts. We randomly select 500 real samples, and for each generator, we pick 500 fake samples. }\label{lpab}
\end{figure}

\begin{table}[t!]
	\centering
	\caption{Gaze-related ablation study. Results are averaged over unseen generators for attribution and detection.} \label{all}
	\setlength{\tabcolsep}{1.9mm}{
		\small
		\begin{tabular}{ccccrr}\toprule\multicolumn{2}{c}{\multirow{3}[6]{*}{Model}} & \multicolumn{2}{c}{Attribution} & \multicolumn{2}{c}{Detection} \\\cmidrule(lr){3-4} \cmidrule(lr){5-6}\multicolumn{2}{c}{} & \multicolumn{2}{c}{Unseen} & \multicolumn{2}{c}{Unseen} \\\cmidrule(lr){3-4} \cmidrule(lr){5-6}\multicolumn{2}{c}{} & \multicolumn{2}{c}{ACC} & \multicolumn{1}{c}{ACC} & \multicolumn{1}{c}{AUC}\\\midrule\multicolumn{2}{l}{GazeCLIP w/o GI} & \multicolumn{2}{c}{74.89} &  85.90     & 70.63\\\multicolumn{2}{l}{GazeCLIP w/o Gaze} & \multicolumn{2}{c}{65.87} &   80.97   &  68.22
			 \\ \multicolumn{2}{l}{GazeCLIP w/ GazePertub.} 
& \multicolumn{2}{c}{59.76} 
& 80.97 
& 63.67 \\
\multicolumn{2}{l}{GazeCLIP w/ GazeMask} 
& \multicolumn{2}{c}{-} 
& 30.04 
& 
46.18 \\
 \multicolumn{2}{l}{GazeCLIP w/ finetuned GE} & \multicolumn{2}{c}{78.48} &    86.39   & 73.71 \\\multicolumn{2}{l}{GazeCLIP (Ours)} & \multicolumn{2}{c}{\textbf{80.66}} &   \textbf{89.80}    & \textbf{77.35} \\\bottomrule\end{tabular}%
	}
	
\end{table}

{\bfseries\setlength\parindent{0em} Impacts of deepfake attribution and detection.} We investigate the interaction between DFA and DFD. We train models to conduct DFA or DFAD, and test their attribution performance on unseen generators. Likewise, we train models to perform DFD or DFAD, and test their detection performance. In Figure~\ref{dfadfd}, due to the introduction of the DFD task, the attribution performance of models tends to decline. Owing to the addition of the DFA task, the detection performance of models is improved. Therefore, attribution promotes detection, while detection may inhibit attribution.

{\bfseries\setlength\parindent{0em} Effects of LoRA number.}
We study the influence of LoRA number in GIE and LRE. We report the performance of GIE from 1 to 4 and LRE from 1 to 2. As Table~\ref{lora} shows, the performance tends to improve with the increase of LoRA number in GIE, when LRE equals 1. The ACC achieves the maximum when two LoRAs are used and starts to decrease thereafter. By contrast, the AUC decreases when two LoRAs in LRE are used. We argue that generally it is easier to mine more common forgery areas when two LoRAs in GIE are used. Too many LoRAs may lead to overfitting to specific forgery traces.

{\bfseries\setlength\parindent{0em} Impacts of gaze injector.}
To delve into the effect of the gaze injector, we conduct the ablation study. In Table~\ref{all}, the attribution ACC of GazeCLIP with GI is about 6\% higher than that of GazeCLIP without GI. We argue that the global gaze prior interaction could allow for more accurate and comprehensive feature alignment and enhanced face forgery patterns. For GI, we employ the class token as the query. Next, we vary the query to study the effect on performance and efficiency. In Table~\ref{query}, we observe that the FLOPs of the model are increased by 1.06G, when all tokens are embraced. Using the class token as a query reduces the computational cost of generating the attention map from quadratic to linear, improving overall efficiency. GazeCLIP with the class token offers an optimal balance between performance and efficiency. As shown in Table~\ref{query}, using all tokens slightly improves attribution ACC, but degrades detection AUC and introduces additional computational overhead. This may be because the gaze embedding represents a global high-level prior rather than spatially aligned local information. Directly injecting the same gaze prior into patch tokens may introduce redundant interactions and disturb local forensic representations. In contrast, the class token aggregates global facial semantics and is therefore better aligned with the global gaze representation. The gaze-enhanced class token can further propagate gaze-aware information to patch tokens through subsequent transformer layers. Consequently, using the class token as the query achieves the best overall trade-off between attribution, detection, and computational efficiency.

{\bfseries\setlength\parindent{0em} Influences of the placement of AWS.} To explore AWS in greater depth, we delve into the effect of the placement of AWS. As Table~\ref{posaws} shows, our method with the AWS placed before MHA and FF in LRE achieves the best performance among models with AWSs at different locations, which underlines that the placement of AWS plays a critical role in maximizing the model effectiveness. We argue that this placement enables the model to create adaptive weight distributions that highlight significant tokens early in the information processing pipeline. As a result, the MHA can operate with a clearer understanding of which words warrant greater attention, leading to more effective aggregations of language features and improved contextual understanding.

\begin{table}[t]
	\centering
	\caption{Region-based attention scores across different unseen generators and datasets.} 	\label{regioncam}
	\setlength{\tabcolsep}{4mm}
	\scriptsize
	\begin{tabular}{c|cccc}
		\toprule
\textbf{Generator} & \textbf{Eye} & \textbf{Mouth} & \textbf{Face} & \textbf{Background} \\
\midrule
SimSwap     & 79.38 & 61.84 & 72.26 & 49.91 \\
InSwapper & 79.79 & 64.34 & 72.95 & 49.66 \\
UniFace     & 79.46 & 62.87 & 72.49 & 49.69 \\
e4s & 81.45 & 65.93 & 74.58 & 50.44 \\
VQGAN & 79.22 & 62.99 & 72.20 & 49.49  \\
SD-2.1      & 75.28 & 38.94 & 67.42 & 46.23 \\
DiT        & 77.88 & 61.73 & 71.17 & 48.87 \\
		SiT         & 78.19 & 61.84 & 71.51 & 48.97 \\
		REFace      & 78.32 & 58.26 & 74.30 & 48.03 \\
	Celeb-DF & 78.97 & 63.68 & 72.25 & 49.47 \\
	DFDC             & 77.14 & 55.17 & 68.55 & 41.67 \\
	WildDeepfake     & 77.15 & 67.41 & 67.94 & 38.16 \\
\midrule
Average          & 78.52 & 60.42 & 71.47 & 47.55 \\
\bottomrule
	\end{tabular}
\vspace{-2em}
\end{table}

{\bfseries\setlength\parindent{0em} Generalizability of gaze as a forgery prior.} 
To further investigate whether gaze itself provides a generalizable and causally meaningful forgery prior for fine-grained deepfake attribution, we conduct both a linear-probing experiment and intervention-based experiments. First, we evaluate whether gaze representations alone are sufficient for forensic prediction. Specifically, face images are fed into the GE and then passed to the detection and attribution expert to generate attribution and detection logits, which are supervised only by the DFA and DFD loss. We train the model using seen generators and CelebA-HQ and test it on the seen and unseen generators. The attribution ACC of GE  is 20.94\% on seen generators, indicating that gaze representations alone are insufficient for distinguishing different generators and are unlikely to capture strong generator-specific shortcuts. The detection ACC and AUC of GE are 90.90\% and 52.03\% on seen generators, respectively. Considering the highly imbalanced real-to-fake ratio, the high detection ACC mainly reflects the majority-class bias, whereas the AUC close to 50\% indicates that gaze representations alone are insufficient for either reliable deepfake attribution or detection. These results show that gaze is not used as a standalone cue for deepfake detection or attribution. Instead, GazeCLIP leverages gaze as a complementary structural prior that interacts with appearance representations to improve forgery representation learning.

Second, to further examine whether gaze makes a causal contribution to the learned forensic representation, we perform three controlled interventions: (1) removing the gaze prior from GazeCLIP, (2) masking the eyeball regions in the input face images while keeping the model unchanged, and (3) perturbing only the gaze embeddings produced by the gaze estimator while keeping the RGB input unchanged. All variants are trained using the same seen generators and CelebA-HQ as the original GazeCLIP and evaluated on unseen generators. In Table~\ref{all}, all three interventions degrade the performance of GazeCLIP on unseen generators. In particular, the attribution ACC of GazeCLIP without gaze decreases by 14.79\%, showing that gaze-related information substantially contributes to cross-generator attribution generalization rather than serving as a redundant auxiliary cue. The attribution ACC of GazeCLIP with perturbed gaze( 59.76\% ) is even lower than that of removing the gaze prior entirely (65.87\%). This result is particularly informative: when informative gaze representations are replaced with corrupted ones, the model performs worse than when no gaze information is provided. This indicates that GazeCLIP actively relies on the semantic integrity of gaze representations: informative gaze cues provide beneficial guidance, whereas corrupted gaze cues mislead the gaze-appearance interaction and degrade cross-generator attribution. The detection AUC of GazeCLIP further drops to 46.18\%, when the eyeball region is masked, which is even lower than that GazeCLIP with perturbed gaze priors (63.67\%). This suggests that directly disrupting the visual source of gaze information has a more severe impact than corrupting the estimated gaze representation alone. 
	
Together, the linear-probing and intervention results provide complementary evidence. The linear-probing experiment shows that gaze alone is insufficient to explain the observed forensic performance, while the intervention experiments demonstrate that removing, corrupting, or masking gaze-related information consistently degrades cross-generator generalization. Thus, gaze contributes meaningfully to generalizable deepfake attribution and detection.

{\bfseries\setlength\parindent{0em} Quantitative analysis.} To examine whether GazeCLIP focuses on transferable facial forgery cues rather than dataset- or generator-specific artifacts, we conduct localization and region-based quantitative analysis. For localization quantitative analyses, we generate class activation maps (CAM) created by GazeCLIP using Grad-CAM for 500 samples per unseen generator and use MediaPipe Face Mesh to define the eye, mouth, face, and background regions. We normalize each attention map and compute the mean CAM activation within the eye, mouth, face, and background regions, with the resulting region-wise attention scores averaged over 500 samples for each generator. A higher score indicates that the model assigns stronger attention to that region during prediction. In Table~\ref{regioncam}, GazeCLIP shows a consistent attention distribution across unseen generators and datasets. The eye region receives the highest average attention (78.52), followed by the face (71.47) and mouth (60.42), while the background receives substantially lower attention (47.55). Importantly, this pattern remains stable across both generators and deepfake datasets, including Celeb-DF, DFDC, and WildDeepfake, suggesting that GazeCLIP primarily attends to transferable facial forensic cues, particularly in the eye region, rather than generator- or dataset-specific background artifacts.

For region-based quantitative analyses, we further construct eye-, mouth-, face-, and background-masked testing sets using MediaPipe Face Mesh landmarks based on the original testing sets, and evaluate the attribution and detection performance of GazeCLIP with and without gaze on unseen generators under these controlled interventions. In Table~\ref{reginmask}, masking the eye, mouth, and background regions increases the attribution accuracy of the gaze-free variant from 65.87\% to 71.71\%, 75.97\%, and 79.38\%, respectively, suggesting that without explicit gaze guidance, the model tends to learn forgery-irrelevant regional appearance patterns rather than stable forensic cues. In contrast, the attribution accuracy of GazeCLIP decreases substantially from 80.66\% to 40.43\% after masking the eye region, while mouth masking leads to a smaller decrease to 60.75\%. These results indicate that GazeCLIP relies more strongly on informative eye-related cues associated with the gaze prior, rather than merely exploiting generic regional appearance variations. Overall, the consistent localization patterns and controlled masking results provide quantitative evidence that GazeCLIP is less prone to dataset- or generator-specific shortcut learning and instead exploits transferable facial forgery cues.

{\bfseries\setlength\parindent{0em} Analysis of gaze-related generalization and potential shortcuts.} To rule out estimator-induced dataset or generator shortcuts, we construct source-matched, head-pose-matched, identity-matched, and image-quality-degraded testing sets based on the original testing set to assess whether GazeCLIP relies on dataset-specific, head-pose, identity, or image-quality artefacts. GazeCLIP with and without gaze are trained using seen generators and CelebA-HQ, and evaluated on unseen generators under these controlled settings. In Table~\ref{tab:control_ablation}, the attribution and detection performance of our model remains relatively stable across different controlled settings, suggesting that GazeCLIP is less sensitive to non-gaze factors such as head pose, eye appearance, identity, image quality, and dataset-specific artifacts. By contrast, the gaze-free variant exhibits larger performance variations across these controlled distributions, suggesting that it is more prone to encoding forgery-irrelevant appearance variations. Moreover, its attribution performance generally improves under the controlled settings, showing that suppressing source-, identity-, pose-, and quality-related nuisance factors reduces manipulation-irrelevant variations and facilitates more reliable forgery attribution. By contrast, GazeCLIP exhibits substantially smaller performance fluctuations, indicating that gaze guidance reduces sensitivity to such nuisance factors and promotes more stable and transferable forgery-related representations.

To further examine whether GazeCLIP simply relies on eye appearance or other region-specific cues, we construct region-masked testing sets based on the original testing sets. As shown in Table~\ref{reginmask}, masking the eye region substantially decreases the attribution accuracy of GazeCLIP from 80.66\% to 40.43\%, while mouth masking causes a smaller drop to 60.75\%. This confirms that eye-related information is important to GazeCLIP. Importantly, the gaze-free variant generally improves under eye-, mouth-, face-, and background-masked settings, suggesting that it tends to encode regional appearance variations as manipulation-irrelevant factors. In contrast, GazeCLIP is more sensitive to removing the eye region, indicating that it exploits informative eye-related cues associated with the gaze prior rather than merely relying on generic eye appearance. Overall, these controlled experiments suggest that the proposed gaze prior does not merely introduce an additional source of overfitting. Instead, it provides a complementary and relatively stable forensic cue that reduces reliance on forgery-irrelevant factors and improves the generalization of models to unseen generators.

\begin{figure}[t!]
	\centering
	\includegraphics[width=\linewidth]{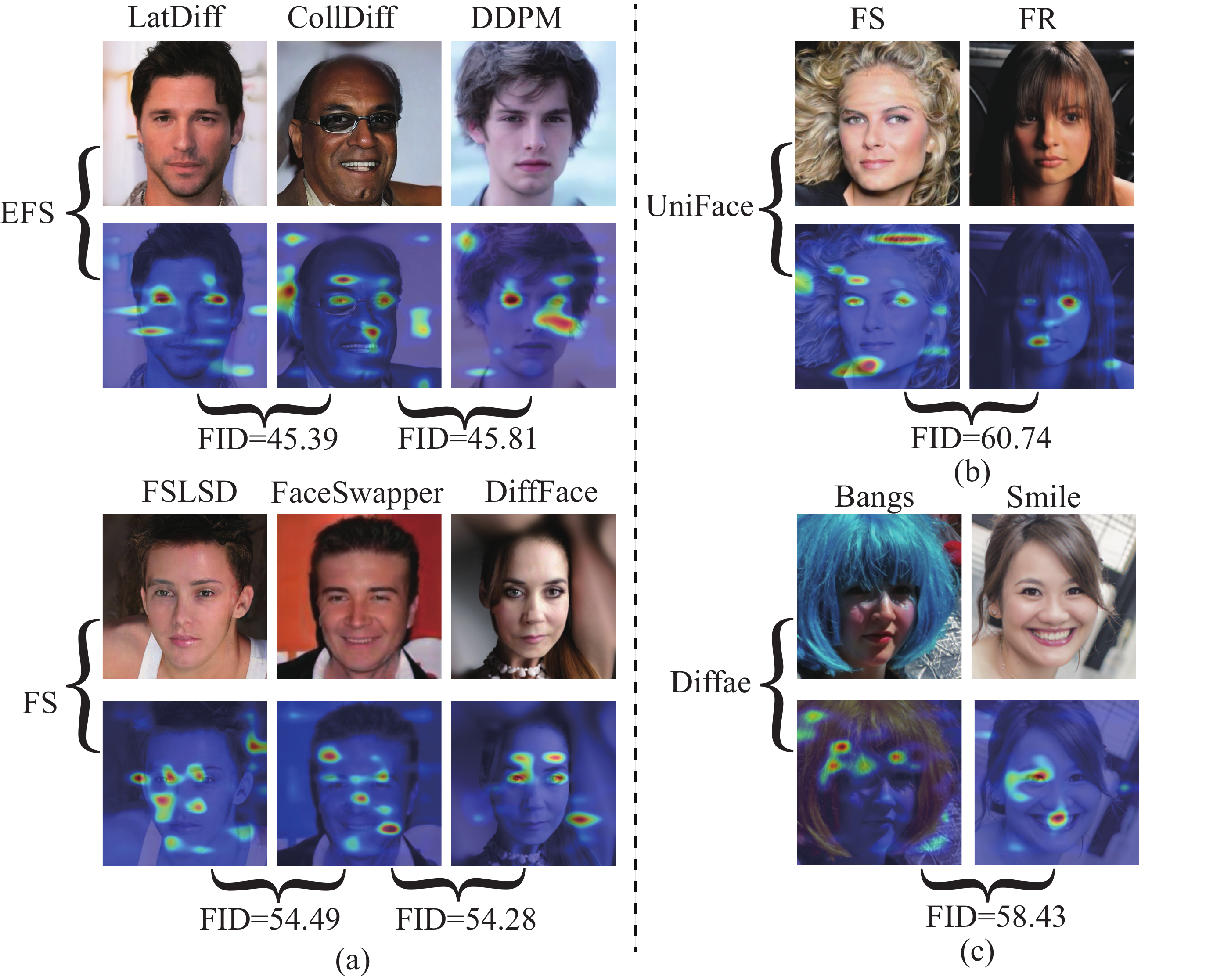}
	\caption{(a) Visualization of heatmaps created by GazeCLIP for various generator samples under the same forgery type, (b) for various forgery type samples generated by the same generator, (c) for various attribute manipulation type samples generated by the same generator.   }
	\label{heat}
\end{figure}

\begin{figure}[t!]
	\centering
	\includegraphics[width=\linewidth]{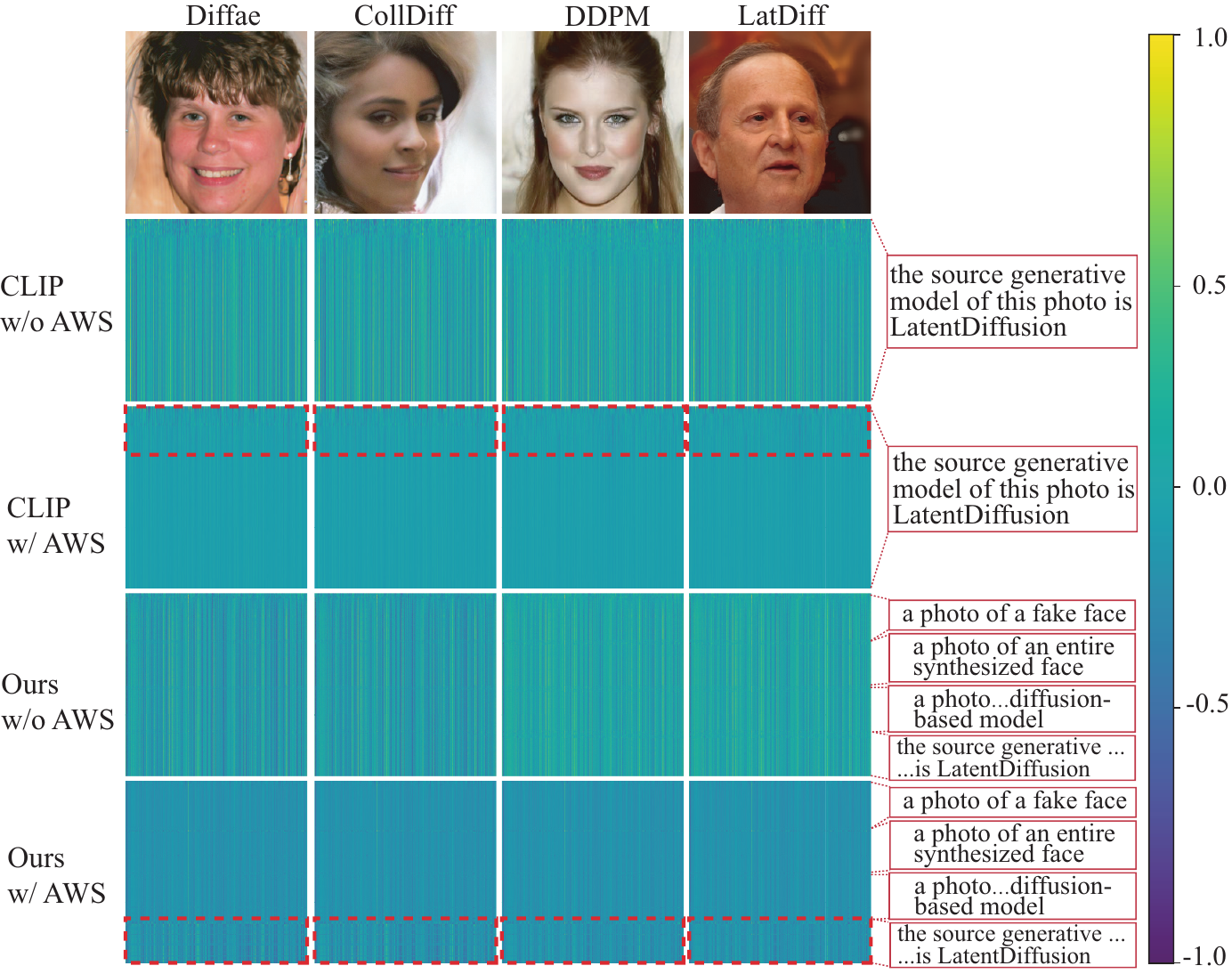}
	\caption{Visualization of language feature matrix created by CLIP or Gaze CLIP (w/o or w/ AWS) for generator samples. The brighter the colour, the more important the word embedding.  }
	\label{dwa}
	\vspace{-2em}
\end{figure}

\section{Visualization}
{\bfseries\setlength\parindent{0em} Visualization of heatmaps.} To study the effect of the gaze prior, we visualize the heatmap created by GazeCLIP with or without gaze using Grad-CAM. In Figure~\ref{Fig4heatmap}, we display the heatmap of various generator samples. Each column displays a face image of the generator. The second to third rows show heatmaps for two models: (a) GazeCLIP without gaze; (b) GazeCLIP with gaze. We notice that (b) explores more long-range gaze-aware forgery details than (a). In Table~\ref{all}, the ACC of GazeCLIP with gaze is around 14.8\% higher than that of GazeCLIP without gaze, showing the value of the gaze prior. To investigate whether GazeCLIP has learned forgery-type-specific attention patterns across different generators, we conducted qualitative and quantitative comparisons of attention map distributions. As Fig.~\ref{heat} (a) shows, we visualize the Grad-CAM heatmaps created by GazeCLIP for each generator under the same forgery type. We randomly select 500 heatmaps for each generator to calculate the FID between their heatmap distributions where a lower FID indicates more consistent attention patterns across generators. In  Fig.~\ref{heat} (b), for face samples belonging to the same forgery type but generated by different models, the FID values are relatively close across generators and lower than those observed when the forgery type varies under the same generator. For example, within the EFS category, the FID between LatDiff and CollDiff and the FID between CollDiff and DDPM are 45.39 and 45.81, respectively. Similarly, for the FS category, the FID values between FSLSD and FaceSwapper and the FID between FaceSwapper and DiffFace are 54.49 and 54.28, respectively. This indicates that GazeCLIP captures consistent forgery cues across different generators within the same forgery type. We further examine whether the attention patterns change when the forgery type varies while the generator is fixed. Following the same protocol, we visualize the Grad-CAM heatmaps for samples generated by the same model under different forgery types and randomly select 500 heatmaps for each forgery type to compute the FID between their attention-map distributions. In  Fig.~\ref{heat} (b), for UniFace-generated face samples belonging to FS and FR, the FID reaches 60.74, which is higher than those observed for samples with the same forgery type but different generators. This larger distribution discrepancy indicates that GazeCLIP adapts its attention patterns to different forgery types. These results suggest that GazeCLIP is more sensitive to forgery-type-related features than to generator-specific artifacts.

We further analyze fine-grained attention differences within the same forgery category. Specifically, for the same AM generator Diffae, we randomly select 500 heatmaps for each attribute manipulation type, i.e., bangs and smile, and compute the FID between their attention-map distributions. As shown in Fig.~\ref{heat}(c), the FID reaches 58.43, indicating that GazeCLIP focuses on different regions for different attribute manipulations generated by the same model. This further suggests that GazeCLIP captures both coarse forgery types and fine-grained manipulation cues, rather than relying solely on generator-specific artifacts.

{\bfseries\setlength\parindent{0em} Visualization of t-SNE.} To investigate the impact of fine-grained language prompts, we visualize the feature distribution created by GazeCLIP (w/o or w/ fine-grained language prompts) using t-SNE \cite{t-SNE} for generator samples. As Figure~\ref{lpab} displays, features of various generators encoded by the GazeCLIP without fine-grained language prompts are clustered in different domains. With the introduction of fine-grained language prompts, the samples of different generators like Diffae and Lattrans from the AM manipulation start to gather together. This proves that fine-grained language guidance could facilitate models to extract more general attribution features.
\section{Conclusion}
In this paper, we build a fine-grained DFAD benchmark and introduce GazeCLIP, a gaze-aware method for guiding CLIP to achieve DFAD. We design VPE to use gaze differences between GAN and diffusion face images to capture gaze-aware forgery features. To realize fully automatic DFAD and release the generalization of CLIP, we propose GIE and LRE. By fusing gaze forgery prompts with general image ones, GIE enables adjusting general image features to be more stable face forgery embeddings, significantly improving the generalization. LRE generates adaptively enhanced fine-grained language prompts via AWS, boosting the creation of an automatic and general DFAD method based on CLIP. Combining VPE, GIE, and LRE, experimental results show that GazeCLIP achieves the state-of-the-art in DFAD.

{\bfseries\setlength\parindent{0em} Limitations and future works.} Manually designing prompts for generators with unknown forgery types is challenging, which may result in limited improvements in model performance. Automatic prompt search or a large language model (LLM) could be employed to generate automated prompts, particularly in zero-shot and few-shot settings, to reduce manual priors and further improve the adaptability and generalization of our framework. Additionally, while the gaze-aware prior significantly boosts generalization, its effectiveness is bounded by the performance of the pre-trained gaze estimator under challenging conditions. Factors such as extreme head poses (e.g., profile views), low resolution (e.g., $<$ 64×64 px), heavy occlusions (e.g., sunglasses or hair), and inaccurate facial alignment may introduce noise into the extracted gaze vectors. Experimental results on distorted images (Fig.~\ref{robust2}) demonstrate that the joint appearance-gaze fusion mechanism mitigates this issue to a large extent, as the appearance encoder and frozen CLIP image prior provide complementary information. In future work, we plan to incorporate a confidence-aware gating module that assigns lower weights to unreliable gaze features and higher weights to appearance-based forgery traces, further enhancing robustness in unconstrained real-world scenarios.

\bibliographystyle{IEEEtran}
\bibliography{GazeCLIP}
\begin{IEEEbiography}[{\includegraphics[width=1in,clip,keepaspectratio]{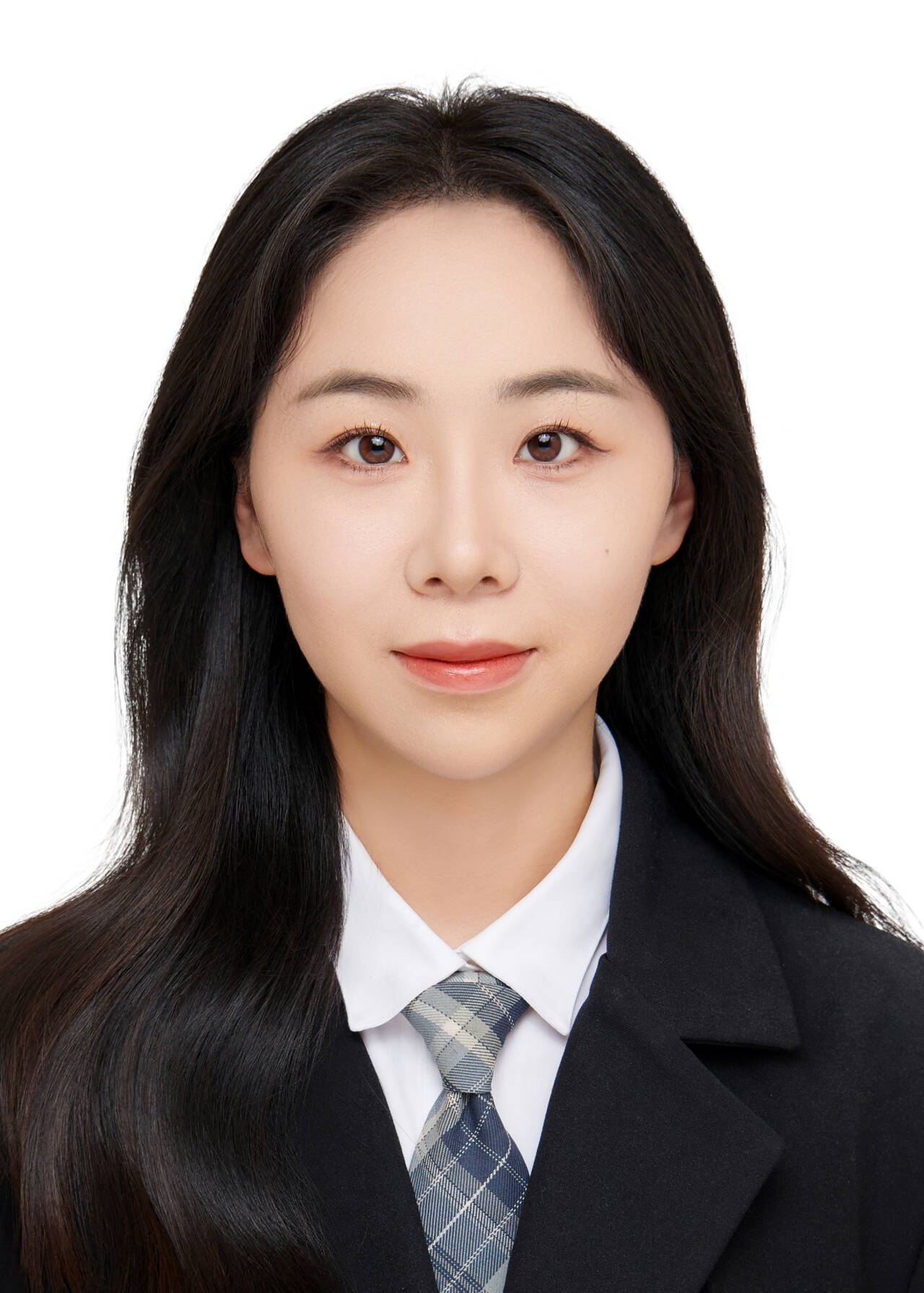}}]{Yaning Zhang} received the double bachelor’s degree in Internet of Things Engineering and English and the M.S. degree in Computer Applied Technology from Qilu University of Technology (Shandong Academy of Sciences), Jinan, China, in 2020 and 2023, respectively, where she is currently pursuing the Ph.D. degree. Her research interests include computer vision, artificial intelligence, multimedia forensics, and face forgery detection.
\end{IEEEbiography}

\begin{IEEEbiography}[{\includegraphics[width=1in,clip,keepaspectratio]{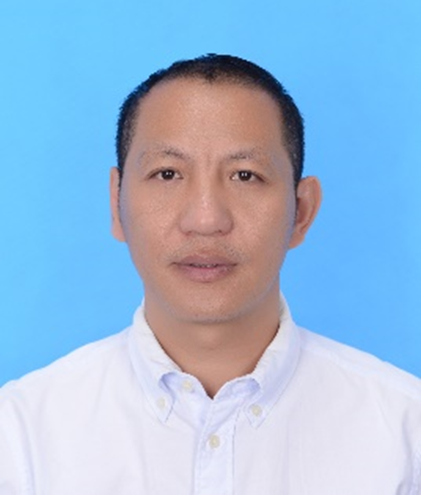}}]{Linlin Shen} (Senior Member, IEEE) is currently a Pengcheng Scholar Distinguished Professor at School of Computer Science and Software Engineering, Shenzhen University, Shenzhen, China. He is also an Honorary professor at School of Computer Science, University of Nottingham, UK. He serves as the Deputy director of National Engineering Lab of Big Data Computing Technology, Director of Computer Vision Institute, AI Research Center for Medical Image Analysis and Diagnosis, and China-UK joint research lab for visual information processing. He also serves as the Co-Editor-in-Chief of the IET journal of Cognitive Computation and Systems and Associate Editor of Expert Systems With Applications. His research interests include deep learning, facial recognition, analysis/synthesis and medical image processing. Prof. Shen is listed as the “Most Cited Chinese Researchers” by Elsevier, and listed in a ranking of the “Top 2\% Scientists in the World” by Stanford University. He received the “Best Paper Runner-up Award” from the journal of IEEE Transactions on Affective Computing, and “Most Cited Paper Award” from the journal of Image and Vision Computing. His cell classification algorithms were the winners of the International Contest on Pattern Recognition Techniques for Indirect Immunofluorescence Images held by ICIP and ICPR. His team has also been the runner-up and second runner-up of a number of competitions for object detection in remote sensing images, nucleus detection in histopathology images and facial expression recognition.
\end{IEEEbiography}

\begin{IEEEbiography}[{\includegraphics[width=1in,clip,keepaspectratio]{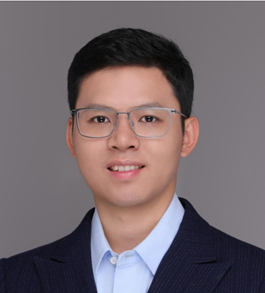}}]{Zitong Yu } (Senior Member, IEEE) received the Ph.D. degree in Computer Science and Engineering from the University of Oulu, Finland, in 2022. Currently, he is an Assistant Professor at Great Bay University, China. He was a Postdoctoral researcher at ROSE Lab, Nanyang Technological University. He was a visiting scholar at TVG, University of Oxford. His research interests include human-centric computer vision and biometric security. He was area chairs of ICME 2023, BMVC 2024, and IJCB 2024. He was a recipient of IAPR Best Student Paper Award, IEEE Finland Section Best Student Conference Paper Award 2020, Best Paper Candidate of ICME 2024, second prize of the IEEE Finland Jt. Chapter SP/CAS Best Paper Award (2022), and World's Top 2\% Scientists 2023 by Stanford University.
\end{IEEEbiography}
\begin{IEEEbiography}[{\includegraphics[width=1in,clip,keepaspectratio]{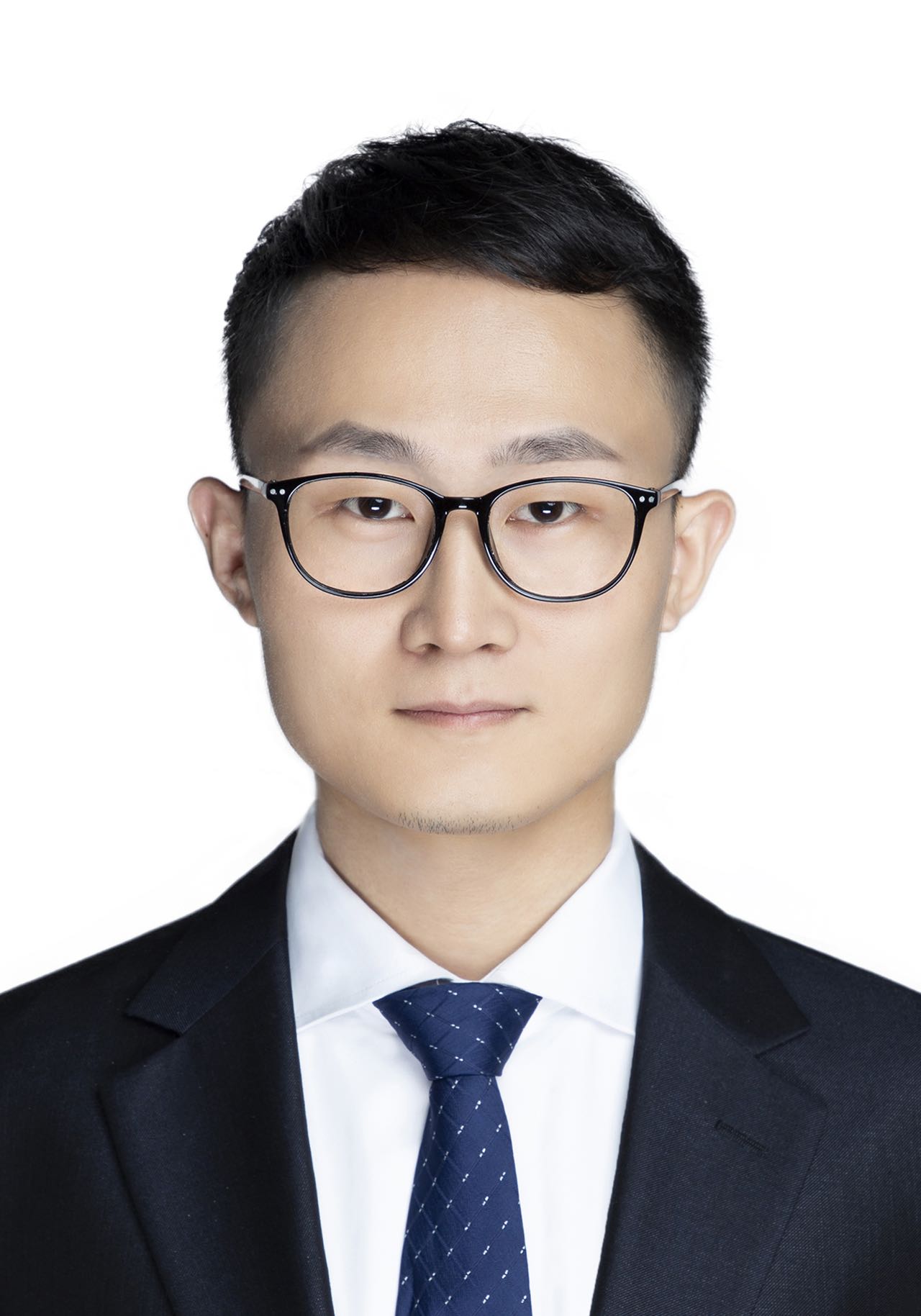}}]{Chunjie Ma} received the Ph.D. degree in Electronic Science and Technology from the Beijing University of Technology in 2023. He is currently a Research Assistant with the Shandong Artificial Intelligence Institute, Qilu University of Technology. His research interests include computer vision, X-ray image prohibited object detection, tamper detection, etc.
\end{IEEEbiography}

\begin{IEEEbiography}[{\includegraphics[width=1in,clip,keepaspectratio]{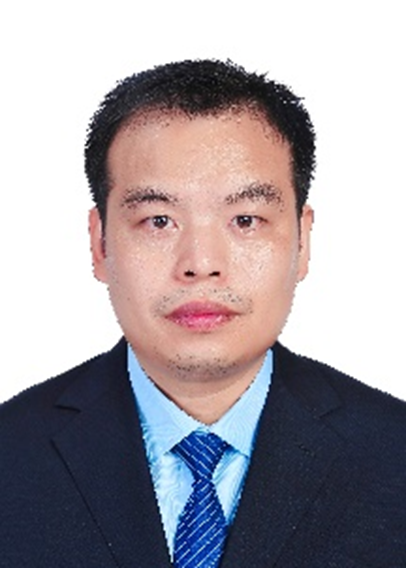}}]{Zan Gao}  (Senior Member, IEEE) received his Ph.D degree from Beijing University of Posts and Telecommunications in 2011. He is currently a full Professor with the Shandong Artificial Intelligence Institute, Qilu University of Technology (Shandong Academy of Sciences). From Sep. 2009 to Sep. 2010, he worded in the School of Computer Science, Carnegie Mellon University, USA. From July 2016 to Jan 2017, he worked in the School of Computing of National University of Singapore. His research interests include artificial intelligence, multimedia analysis and retrieval, and machine learning. He has authored over 100 scientific papers in international conferences and journals including TPAMI, TIP, TNNLS, TMM, TCYBE, CVPR, ACM MM, WWW, SIGIR and AAAI.
\end{IEEEbiography}

\end{document}